\RequirePackage{fix-cm}
\documentclass[twocolumn]{svjour3}          
\smartqed  
\usepackage{graphicx}
\usepackage{hyperref}
\hypersetup{
    colorlinks=true,
    breaklinks=true,
    linkcolor=red,
    filecolor=magenta,      
    urlcolor=blue,
    citecolor=blue
}
\usepackage{pifont}
\usepackage{multirow}
\usepackage[caption=false]{subfig}
\usepackage{amsmath}
\usepackage{rotating}
\usepackage[dvipsnames]{xcolor}
\usepackage{comment}

\newcommand{\cmark}{\ding{51}}
\newcommand{\xmark}{\ding{55}}

\definecolor{best_result}{RGB}{0,0,200}
\definecolor{bf_best_result}{RGB}{0,0,200}
\definecolor{updated}{RGB}{200,0,0}

\begin{document}

\title{Robust Text Line Detection in Historical Documents: \\ Learning and Evaluation Methods
}


\author{Mélodie Boillet         \and
        Christopher Kermorvant  \and
        Thierry Paquet
}


\institute{M. Boillet \at
              TEKLIA, Paris, France \\
              LITIS, Normandie University, Rouen, France \\
              \email{boillet@teklia.com} 
           \and
           C. Kermorvant \at
              TEKLIA, Paris, France \\
              LITIS, Normandie University, Rouen, France \\
              \email{kermorvant@teklia.com} 
           \and
           T. Paquet \at
              LITIS, Normandie University, Rouen, France \\
}

\date{Received: date / Accepted: date}

\maketitle

\begin{abstract}

Text line segmentation is one of the key steps in historical document understanding. It is challenging due to the variety of fonts, contents, writing styles and the quality of documents that have degraded through the years.

In this paper, we address the limitations that currently prevent people from building line segmentation models with a high generalization capacity. We present a study conducted using three state-of-the-art systems Doc-UFCN, dhSegment and ARU-Net and show that it is possible to build generic models trained on a wide variety of historical document datasets that can correctly segment diverse unseen pages. This paper also highlights the importance of the annotations used during training: each existing dataset is annotated differently. We present a unification of the annotations and show its positive impact on the final text recognition results. In this end, we present a complete evaluation strategy using standard pixel-level metrics, object-level ones and introducing goal-oriented metrics.


\end{abstract}

\section{Introduction}
\label{introduction}
In the past few decades, a real interest in understanding and analyzing images of historical documents has emerged, mostly motivated by Human and Social Sciences research. Meanwhile, with the development of deep learning techniques, the automatic processing of such documents has become more and more effective, making handwriting recognition a reality. Therefore, many systems and strategies have been proposed to deal with a wide variety of tasks such as the detection and recognition of names (persons and places), dates \cite{michael2019} or the detection of signatures \cite{tarride}.



Since one of the final goals in analyzing images of historical documents is to understand the texts, the systems developed so far rely on two key components: Handwritten text line detection and Handwritten Text line Recognition (HTR). Some other pioneer contributions have also investigated line segmentation-free approaches that try to directly recognize the text from paragraphs or pages \cite{origaminet} \cite{bluche16}; however, these systems still face important limitations and struggle to be effective on complex documents. \\

The literature shows that competitive and robust systems have been developed to tackle the text line detection problem, showing good performances on individual datasets. However, they often perform poorly on other unseen documents, lacking in generalization capacity.
We advocate that the main reasons for this limitation are twofold:

\begin{enumerate}
    \item The annotation schemes are not always compatible, making the combination of datasets in a single unified training set difficult. These different schemes induce a label bias of the datasets and do not allow a fair comparison of segmentation approaches from one dataset to another. Therefore, no comparison of systems has been conducted on a large and diverse dataset, both in training and testing.
    \item The evaluation metrics at pixel-level are not always representative of the real impact of the segmentation stage on the recognition stage and thus of the generalization capacities. In addition, comparisons of text line segmentation systems in terms of recognition error rates are rarely reported due to the complexity of this evaluation. \\
\end{enumerate}

In the literature, handwritten text line detection studies usually focus on developing a specific neural network architecture as well as a good training strategy. However, they often omit to analyze the annotations used during training and testing, whereas they are as important as the network architecture itself, since they guide the training phase and the final results. Indeed, the label bias is particularly crucial when one wants to analyze the impact of the segmentation stage on the recognition stage, but it is not considered at all in almost every study focusing on text line segmentation. It is neither considered by studies focusing on handwriting recognition, as the segmentation process falls out of their scope. For instance, the first letter in historical documents is often adorned, adding it or not in the text lines during the annotation process can really impact the final recognition results, hence the importance of carefully creating and analyzing the annotations. Another common problem arises with text line segmentation when two annotated bounding polygons are touching. In such a situation, the network generally provides merged lines that will not be recognized by the HTR system, while generally the evaluation metrics do not account for these problems. Indeed, the very popular and common Intersection-over-Union (IoU) metric can not detect these situations and to fairly count the correct and wrong line splits. In this paper, we address this problem by introducing a unifying labeling strategy of the datasets which reduces the bounding polygons overlaps to get coherent output with the required input of the HTR systems.

A second important point in processing documents is the evaluation process. Most of the existing line segmentation methods are evaluated and compared with pixel-level metrics such as the IoU, Precision, Recall and F1-score. Even if these metrics indicate how the network performs, they give no actual information about the amount of information retrieved (number of detected lines). Some object-level metrics like the Average Precision (AP) have been proposed to overcome this problem; however, they always rely on a fixed IoU threshold. In this paper, we analyze these limitations and introduce the mean threshold-free mAP metric already used in the COCO Detection Challenges. Furthermore, as handwritten line detection is the first step of the whole recognition process, it should be more realistic to evaluate its true impact on the final recognition results (Character and Word Error Rates), by conducting a goal-directed evaluation \cite{trier}. In this respect, we also propose an evaluation strategy based on the recognition results obtained after a HTR system. \\

In this paper, we provide a fair and extensive evaluation of three state-of-the-art approaches for text line segmentation, Doc-UFCN \cite{boillet2020}, dhSegment \cite{dhsegment} and ARU-Net \cite{gruning2018} on a large collection of historical datasets and using multiple metrics including a goal-directed one. We analyze the line segmentation metrics of the literature regarding multiple datasets publicly available and show some label inconsistencies among the datasets. Our first contribution is to propose a unifying labeling strategy of the datasets that prevent from the label bias for the line detection task. This modified labeling jointly allows encompassing the variability in the annotations and to train models with a higher generalization capacity.
Second, we conduct an evaluation of the state-of-the-art using different metrics and training protocols, allowing to build more generic segmentation models that encompass the afford-mentioned limitations and achieve similar, if not better, results than the models trained on single datasets. We show that the obtained models can correctly segment pages from unseen datasets and even outperform specific models on these pages. In the end, we demonstrate that the proposed unified labeling has a positive impact on the segmentation process and HTR performance.

This paper is organized as follows: Section \ref{sec:related_works} presents a literature review of recent approaches developed to carry out HTR with or without explicit text-line segmentation. We also review the current evaluation methods. In Section \ref{sec:dataset}, we introduce the datasets used to conduct our study, as well as our unifying labeling strategy. Section \ref{sec:training_details} briefly recalls the three state-of-the-art line segmentation approaches selected to conduct the experimentations and their training details. Lastly, Sections \ref{sec:seg_eval} and \ref{sec:gd_eval} present segmentation and goal-directed evaluations of the different experiments.

\section{Related work}
\label{sec:related_works}
In this section, we review the main approaches for image segmentation applied to text line detection in documents and describe the different metrics used for the evaluation of document segmentation systems.

\subsection{Line segmentation-free and end-to-end systems}
\label{sec:rw_end2end}

Few recent segmentation-free systems try to segment paragraphs or pages and recognize their texts at once. For example, Bluche \cite{bluche16} proposed a segmentation-free handwriting recognition system based on MDLSTM layers. They apply successive MDLSTM and convolutional layers on the 2D input image and get the character probabilities with a final fully connected layer followed by a softmax normalization. Yousef \textit{et al.} \cite{origaminet} recently proposed OrigamiNet, a segmentation-free recognition system. Their strategy is to run up-scaling operations on the 2D input feature map to unfold a paragraph into a single line that contains all the input image lines. The system is trained with a standard CTC loss, and has shown promising results on the IAM \cite{iam} and ICDAR2017 HTR \cite{icdar2017_htr} datasets.

The advantage of these line segmentation-free systems is that they require fewer annotations (only the textual transcriptions) compared to systems that rely on line segmentation. Moreover, they can achieve better recognition results as they are not impacted by line segmentation errors that are introduced in two-stage approaches. However, these networks are still limited to simple layout images. So far, they cannot deal with two-columns document images. In addition, training such networks can be computationally prohibitive on very large datasets. For the time being, these architectures are not suitable to deal with complex layout and diverse historical document datasets. \\

End-to-end systems have also been proposed to run both text localization and recognition using a single training framework. Deep TextSpotter \cite{deepTextSpotter} uses a modified version of YOLOv2 \cite{yolov2} for the detection and a CNN for the recognition of natural scene text images. Both models are trained simultaneously, which improves the performances compared to standard methods. Wigington \textit{et al.} \cite{wigington2018} proposed to jointly train a network to detect the start of text lines, a line follower and a final recognition network. This joint training allows the three sub-models to mutually adapt and supervise each other. Despite good results, this method needs pre-training on additional segmented images. 

\subsection{Image segmentation}
\label{sec:rw_image_seg}

In the computer vision field, the object detection literature can be split into three main categories: region proposal, bounding boxes regression and pixel-level segmentation—based systems.

Region proposal-based systems consist in three consecutive steps. First, a bunch of category-independent region proposals (bounding boxes) is generated. Then a Convolutional Neural Network (CNN) is applied over these regions to extract the meaningful information and a classifier predicts the class of each region proposal. This strategy has been first proposed by Girshick \cite{R-CNN} and applied to natural scene images of the VOC 2010-2012 datasets. Despite the significant improvement in the field and the development of more advanced systems (Fast R-CNN \cite{fast_rcnn}, Faster-RCNN \cite{faster_rcnn} and \cite{zhong2017}), this method did not break through the document images community. Indeed, they are well suited for natural scene images where only few objects are present on the images, contrarily to document images. \\

Bounding boxes regression systems have been first introduced by Redmon \textit{et al.} with their YOLO model \cite{yolo}. The goal of YOLO was to be end-to-end and faster than R-CNN methods also on natural scene images. The image is first divided into a grid, then each grid cell predicts a pre-defined number of bounding boxes with their confidences as well as class probabilities using a single neural network. The final detections are the boxes with the highest confidence score and the highest class probability in this box. Similar methods \cite{yolov2} \cite{yolov3} have emerged following this idea; however, very few were applied to document images, probably for the same reason mentioned above: document images contain too many objects to be detected.

Moysset \textit{et al.} \cite{moysset16} used regressions for detecting text lines in documents. Based on the idea of YOLO, bounding boxes and confidence scores are predicted using a CNN. They tried two regression strategies: directly predicting the bounding boxes and predicting only the bottom left and top right points before pairing them. The second strategy has shown a real improvement for the box detection task on Maurdor \cite{maurdor} documents.

\subsection{Pixel-level segmentation}
\label{sec:rw_pixel_seg}

Mainly coming from the biomedical domain and popularized by Ronneberger \textit{et al.} with their U-Net architecture \cite{unet}, the goal of this kind of networks is to output pixel-wise class probabilities. In these last years, many systems have been developed to deal with pixel-level document image segmentation. Some are focusing only on text line detection, whereas some others can deal with different tasks and multiple classes.

To solve the text line detection task, most of the existing systems are using Fully Convolutional Neural Networks (FCN). Barakat \textit{et al.} \cite{barakat2018} presented a U-shaped architecture composed of convolutions and pooling layers in the encoding step, followed by upsampling layers in the decoding stage. Unlike the original U-Net \cite{unet}, they do not use high-level feature maps during the decoding stage, but only low-level ones to make the final predictions. Following the same idea, Mechi \cite{mechi2019} proposed an adaptive U-Net architecture, quite similar to the first one, with convolutions in the encoder and transposed convolutions in the decoder. However, they use high-level feature maps during the decoding stage. Both models have shown good results on historical datasets; however, the first one requires binarized input images. Renton \textit{et al.} \cite{renton2018} went further with this U-Net architecture by replacing the standard convolutions of the encoder by dilated ones. This allows the network to have a larger receptive field and to use more context. Their model provided second best results on the well-known cBAD dataset \cite{cBAD2017} while having a really light post-processing compared to the first best model. Lastly, in one of our previous works we presented Doc-UFCN \cite{boillet2020}, a FCN inspired by the previously cited systems and by the multi-modal model presented by Yang \textit{et al.} \cite{yang2018}. The encoder is composed of dilated convolutions, while the decoder consists in transposed convolutions. This system has show good performances on various historical documents while using fewer parameters and requiring less computing for the inference.

For the baseline detection task, Grüning \textit{et al.} \cite{gruning2018} proposed ARU-Net, an extended version of the U-Net architecture using spatial attention to deal with various font sizes and residual blocks to reach higher results. In addition, they apply a complex post-processing step to generate baselines from the pixel-level predictions. \\

\sloppy
Systems dealing with multiple classes have also been proposed. Oliveira \textit{et al} \cite{dhsegment} proposed dhSegment, a generic system for historical document processing. Their model follows an encoder-decoder architecture where the encoder is a ResNet-50 \cite{resnet} pre-trained on natural scene images such as ImageNet \cite{ImageNet}. Even if this method has shown good results on various tasks (page extraction, baseline detection or layout analysis), the inference time is still significant as shown in one of our early works \cite{boillet2020}. Yang \textit{et al.} \cite{yang2018} also presented a model for multiple classes segmentation. As for the aforementioned systems, they designed a U-shaped FCN model applied to modern documents. Their model is able to detect various classes of text (section headings, lists, or paragraphs) and figures using both text and visual contents. This method has shown good performances on modern documents, measured by the Intersection-over-Union metric.

\begin{table*}[ht]
    \caption{Pixel and object-level metrics used by recent systems for document object detection.}
    \label{tab:metrics}
    \begin{center}
        \begin{tabular}{l|ccc|cccc}
            \hline\noalign{\smallskip}
            \multicolumn{1}{c}{\multirow{2}{*}{}} & \multicolumn{3}{c}{\textsc{Pixel-level}} & \multicolumn{4}{c}{\textsc{Object-level}} \\
            \multicolumn{1}{c}{} & \multicolumn{1}{c}{IoU} & \multicolumn{1}{c}{P/R} & \multicolumn{1}{c}{F1} & \multicolumn{1}{c}{P/R} & \multicolumn{1}{c}{R@.85/.95} & \multicolumn{1}{c}{mAP@.65} & \multicolumn{1}{c}{mAP} \\ \noalign{\smallskip}\hline\noalign{\smallskip}
            Barakat \cite{barakat2018} & & \cmark & \cmark & & & & \\
            Mechi \cite{mechi2019} & \cmark & \cmark & \cmark & & & & \\
            Renton \cite{renton2018} & & \cmark & \cmark & & & & \\
            Doc-UFCN \cite{boillet2020} & \cmark & \cmark & \cmark & & & & \\
            dhSegment \cite{dhsegment} & \cmark & \cmark & \cmark & & \cmark & & \\
            Yang \cite{yang2018} & \cmark & & \cmark & & & & \\
            Tarride \cite{tarride} & & \cmark & \cmark & \cmark & & \cmark & \\
            Soullard \cite{soullard2020} & & & & & & & \cmark \\
            Melnikov \cite{melnikov20} & & \cmark & \cmark & & & \\
            \noalign{\smallskip}
            \hline
        \end{tabular}
    \end{center}
\end{table*}

\subsection{Evaluation methodology}

\fussy
Assessing the quality of detection or recognition algorithms requires using the best appropriate metrics. However, one also has to focus on the annotated data used during training and testing. If the data annotations are inconsistent with the metric at end, then the metric cannot reflect the real performances of the model. This problem has been little studied in the literature. For example, when confronted to touching bounding boxes, Melnikov and Zagaynov \cite{melnikov20} suggested removing the ascenders and descenders by reducing the height of the annotated boxes by 30\% at the top and the bottom. Then, they downscaled the rasterized polygons to the input resolution to train their model. Even if this method has proved to be effective to reduce the label bias of the annotation, some problems remain when dealing with vertical and inclined lines.

In the same spirit, Peskin \textit{et al.} \cite{Peskin20} have proposed various annotation masks for detecting and classifying geometric shapes from gray-scale images. They suggested labelling the objects (circles, rectangles, and triangles) in four ways: with small center marks, large center marks, small center marks with a 1-pixel outline of the object and small center mark with a 2-pixels outline of the object. As far as localization problems are concerned, they have shown that small center marks yield the best results. However, better classification results are obtained using the small center marks with outlines. As we can see, finding the most suited annotation is not a trivial problem. \\

In addition to annotation problems, one has to focus on finding appropriate metrics to properly evaluate and compare segmentation results. Hemery \textit{et al.} \cite{metrics_study} have studied the key properties a metric should have for a localization task. From the analysis of 33 existing metrics, they have established the most suitable ones for that task. Following this idea, we show in this paper that the currently main used metrics are not sufficient for evaluating and comparing line segmentation models.
\sloppy
As shown in Table \ref{tab:metrics}, the majority of existing pixel-level segmentation systems are evaluated using pixel-level metrics. The Precision and Recall measures are widely used, as well as the Intersection-over-Union (IoU) and F1-score. They are based on the amount of correctly predicted pixels. However, these metrics provide no information about the number of correctly predicted and missed or split objects. Computing these values at line/object-level is not directly applicable, since the decision whether an object has been detected or not is more difficult than for a pixel.

\fussy
To tackle this problem, metrics originally designed in the Information Retrieval community \cite{perf_evaluation_IR} have been adapted to images and used during the PASCAL VOC Challenge 2012 to compute the Precision at object-level. During this competition, the detection task was evaluated based on the Precision-Recall curve at object-level, where the detections are considered as true or false positives according to their area of overlap with the ground-truth objects. According to this approach, Tarride \textit{et al.} \cite{tarride} first associate the predicted objects with those of the ground-truth and consider a prediction as true positive if its IoU is higher than a chosen threshold \textit{t = 0.65}. That way, they can compute the Precision (P@.65), Recall (R@.65) and mean Average Precision (mAP@.65) at object-level.

Wolf and Jolion \cite{wolf06} have shown the importance of both the detection quality (accuracy of the detected objects) and the detection quantity (number of objects) when evaluating a system. The mAP measure, which is the area under the Precision-Recall curve, allows evaluating the detection quantity based on a given quality criterion: the IoU threshold. To be able to measure both detection quality and quantity, the use of mAP averaged over a range of IoU thresholds have emerged for the detection of objects. Soullard \textit{et al.} \cite{soullard2020} used this mean mAP for evaluating their historical newspaper segmentation model.

The ZoneMap metric proposed by Galibert \textit{et al.} \cite{zonemap} also evaluates document segmentation systems at object-level and does not rely on any threshold. It is based on links between hypothesis and reference zones. First, the forces of the links are computed: if a predicted zone is correct, then the force with a reference zone will be high. On the contrary, all forces for a false positive zone will be low. Then, zones are grouped according to these links and each group receives a segmentation and a classification error, which are computed based on the type of group (Match, Miss, False Alarm, Merge or Split). These two errors are further combined to give one single value. Even if this metric has shown to be complementary to the IoU metric on Maurdor evaluation \cite{maurdor}, it is at present not really used due to the complexity of its computations and its hard applicability to images with multiple objects.

Lastly, Trier and Jain \cite{trier} have shown the importance of a goal-oriented evaluation for binarization methods, since an evaluation from a human expert really depends on his visual criteria. They applied 11 of the most promising locally adaptive binarization methods to test images before feeding the results to an OCR module. The binarization methods were then compared using the recognition, error and reject rates. Similarly, Wolf and Jolion \cite{wolf06} have shown the importance of these measures for object detection. Indeed, for their task, final users want to know whether the objects have been detected or not, no matter the exact quantity of detected pixels. This is what one of our proposals is aiming at: using goal-directed HTR metrics to evaluate text segmentation systems. To the best of our knowledge, there are no previous work in the literature doing this.

\begin{table*}[ht]
    \caption{Details of the datasets: whether it is public or not, number of manuscripts from which the pages have been extracted, number of images and lines, presence of transcriptions, mean number of characters by line and mean size of the images. The character densities and mean sizes have been computed on the training sets except for the two testing datasets ScribbleLens and Alcar.}
    \label{tab:datasets}
    \begin{center}
        \setlength\tabcolsep{5pt}
        \begin{tabular}{l|cr|rrr|rrr|cr|c}
            \hline\noalign{\smallskip}
            \multicolumn{1}{c}{\multirow{2}{*}{\textsc{Dataset}}} & \multicolumn{1}{c}{\multirow{2}{*}{\textsc{Public}}} & \multicolumn{1}{c}{\textsc{Manus. /}} & \multicolumn{3}{c}{\textsc{Images}}
            & \multicolumn{3}{c}{\textsc{Lines}} & \multicolumn{1}{c}{\multirow{2}{*}{\textsc{Text}}} & \multicolumn{1}{c}{\multirow{2}{*}{\textsc{Density}}} & \multicolumn{1}{c}{\multirow{2}{*}{\textsc{Mean size}}} \\
            \multicolumn{1}{c}{} & \multicolumn{1}{c}{} & \multicolumn{1}{c}{\textsc{Charters}} & \multicolumn{1}{c}{train} & \multicolumn{1}{c}{dev} & \multicolumn{1}{c}{test} & \multicolumn{1}{c}{train} & \multicolumn{1}{c}{dev} & \multicolumn{1}{c}{test} & \multicolumn{1}{c}{} & \multicolumn{1}{c}{} & \\ \noalign{\smallskip}\hline\noalign{\smallskip}
            AN-Index & \xmark & N/A & 19 & 3 & 12 & 433 & 62 & 171 & \xmark & N/A & 2078$\times$1433 \\
            Balsac \cite{Vezina2020} & \xmark & 74 & 730 & 92 & 91 & 36807 & 4577 & 4301 & \cmark & 26 & 3882$\times$2418 \\
            BNPP & \xmark & 5 & 7 & 2 & 3 & 705 & 218 & 358 & \cmark & 34 & 3723$\times$5040 \\
            Bozen \cite{bozen} & \cmark & N/A & 350 & 50 & 50 & 8367 & 1043 & 1140 & \cmark & 20 & 3511$\times$2394 \\
            cBAD2019 \cite{cBAD2019} & \cmark & 7 & 755 & 755 & 1511 & 49630 & 46724 & 209714 & \xmark & N/A & 2980$\times$2569 \\
            DIVA-HisDB \cite{diva} & \cmark & 3 & 60 & 30 & 30 & 6037 & 2999 & 2897 & \xmark & N/A & 4992$\times$3328 \\
            HOME-NACR \cite{boros2020} & \cmark & 43 000 & 398 & 49 & 49 & 7004 & 900 & 840 & \cmark & 138 & 4635$\times$6593 \\
            Horae \cite{horae2019} & \cmark & 500 & 522 & 20 & 30 & 12568 & 270 & 958 & \cmark & 28 & 4096$\times$5236 \\
            READ-Complex \cite{gruning_cbad} & \cmark & \multirow{2}{*}{9} & 216 & 27 & 27 & 17768 & 2160 & 1758 & \xmark & N/A & 4096$\times$2733 \\
            READ-Simple \cite{gruning_cbad} & \cmark & & 172 & 22 & 22 & 5117 & 539 & 723 & \xmark & N/A & 3947$\times$2721 \\ \noalign{\smallskip}
            \hline\noalign{\smallskip}
            ScribbleLens \cite{scribblelens} & \cmark & N/A & - & - & 21 & - & - & 563 & \cmark & 52 & 3793$\times$2558 \\
            Alcar \cite{Stutzmann2021} & \cmark & 17 & - & - & 55 & - & - & 2727 & \cmark & 108 & 4080$\times$4853 \\ \noalign{\smallskip}
            \hline
        \end{tabular}
    \end{center}
\end{table*}

\section{Historical document datasets}
\label{sec:dataset}
One of our goals being to develop a generic text line detector on historical documents, we collected 9 historical datasets including 6 public ones to conduct the experimentations. A description of these datasets is given in Table \ref{tab:datasets}. In addition, since we want to evaluate the generalization capacity of the obtained models, we collected two additional datasets ScribbleLens \cite{scribblelens} and Alcar \cite{Stutzmann2021} that are used only during the evaluation step.

\subsection{Datasets}

\paragraph{AN-Index}: This first dataset is composed of 34 document images of digitized finding aid from the French National Archives. The documents are written in French.

\paragraph{Balsac}: This second dataset \cite{Vezina2020} consists in 913 images of birth, marriage and death records extracted from 74 Québécois registers.
²
\paragraph{BNPP}: This dataset was provided by the Historical Archives\footnote{\url{https://history.bnpparibas/}} of the BNP Paribas bank. It consists of a sample of 12 images extracted from 5 scanned registers of minutes from the Comptoir National d'Escompte of Paris. They have been selected among a hundred registers written in French between the 19th and 20th century.

\paragraph{Bozen}: This dataset \cite{bozen} is part of the READ project and consists in 450 annotated handwritten pages. The pages are extracted from documents of the Ratsprotokolle collection, written between 1470 and 1805.

\paragraph{cBAD2019}: The cBAD dataset \cite{cBAD2019} consists in 3021 annotated document images collected from seven European archives. It has been used during the cBAD competition at ICDAR2019 for baseline detection.

\sloppy
\paragraph{DIVA-HisDB}: DIVA-HisDB \cite{diva} is a well-known dataset that is composed of 120 annotated handwritten pages extracted from 3 medieval manuscripts.

\fussy
\paragraph{HOME}: The HOME dataset \cite{boros2020} consists in 420 annotated medieval charters selected among 43,000 digitized charters from the archives of the Bohemian Crown and the archives of monasteries. They have been written from 1145 and 1491 in Early Modern German, Latin, and Czech.

\paragraph{Horae}: Horae dataset \cite{horae2019} consists in 573 annotated pages from 500 different books of hours dating from the Middle Ages.

\paragraph{READ-BAD}: This dataset \cite{gruning_cbad} is composed of 2036 archival document images written between 1470 and 1930. The pages have been extracted from 9 archives and split into two subsets: Simple and Complex, depending on the complexity of their layout. Both subsets have been used during the cBAD competition at ICDAR2017 \cite{cBAD2017} for the baseline detection task.

\paragraph{ScribbleLens}: ScribbleLens \cite{scribblelens} contains 1000 pages of Early Modern Dutch documents such as ship journals and daily logbooks produced between the 16th and 18th century. The manuscripts consist in journeys written by captains and traders from the Vereenigde Oost-indische Company (VOC). The test set is composed of 21 pages annotated and transcribed at line-level.

\sloppy
\paragraph{Alcar}: The HOME IRHT Alcar dataset \cite{Stutzmann2021} contains 17 cartularies produced between the 12th and 14th century and written in Latin. To generate a test set, we collected 55 pages annotated at line-level with their corresponding transcriptions among all the manuscripts.  \\

\begin{figure*}
    \centering
    \includegraphics[width=0.95\textwidth]{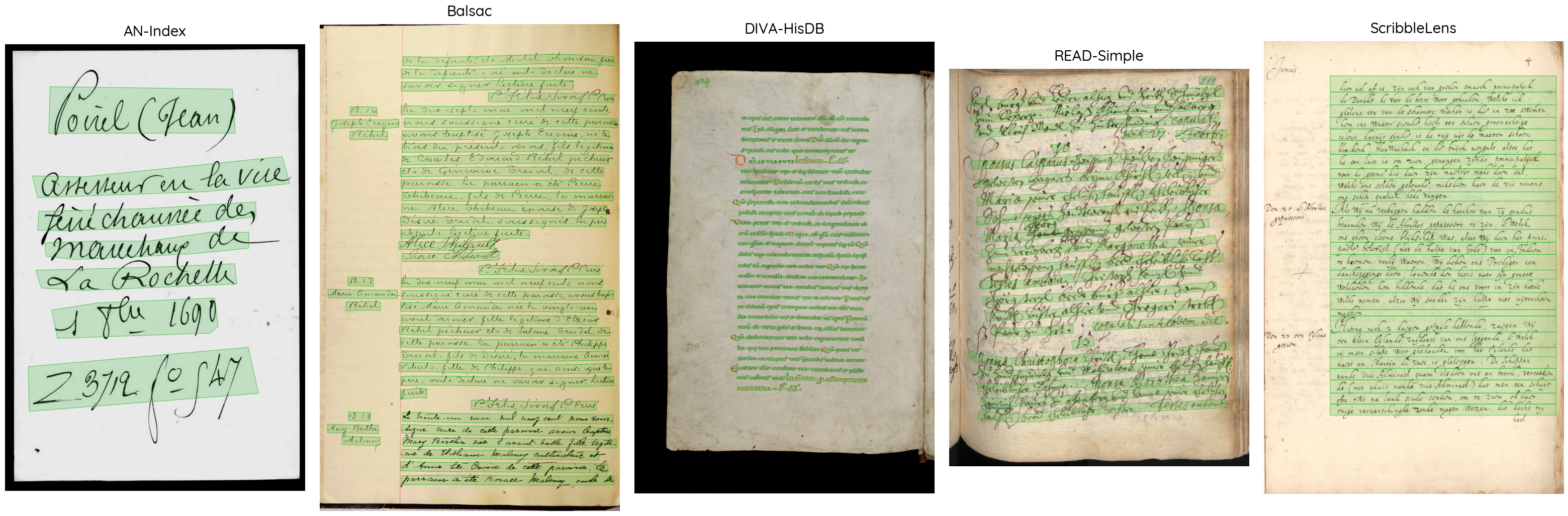}
    \includegraphics[width=0.95\textwidth]{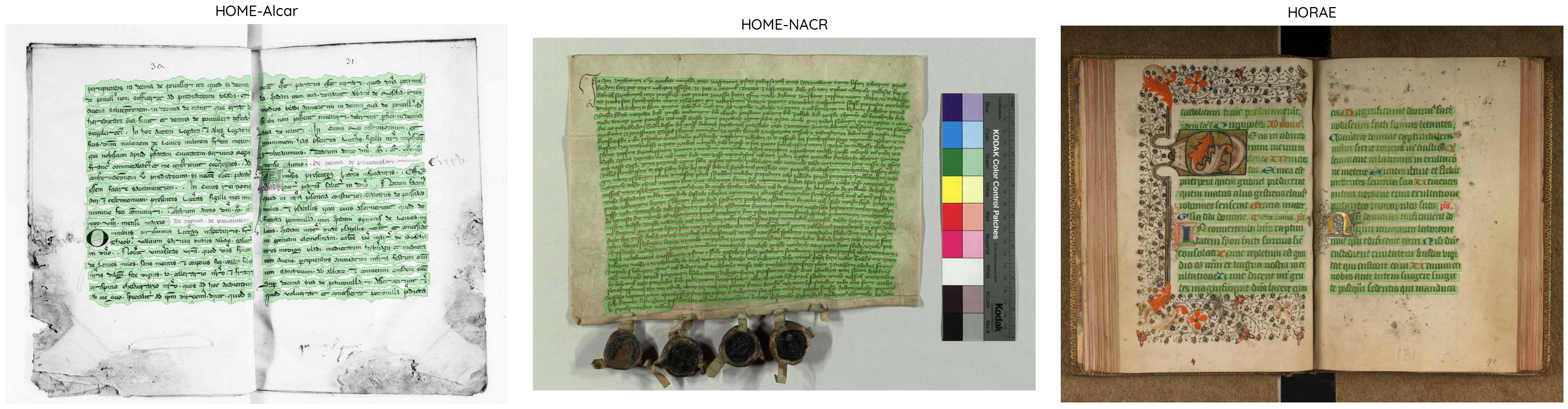}
    \caption{Example of dataset images with their annotations.}
    \label{fig:datasets}
\end{figure*}

\fussy
All these datasets have been chosen for their diversity in sizes, writings, and layouts. Figure \ref{fig:datasets} presents the documents variety by showing some example images with their line annotations. The split used to train the models has been obtained by simply grouping the respective training and validation subsets. \\

\subsection{Labels analysis}

All the datasets cited above have been used to train generic text line detection models using Doc-UFCN \cite{boillet2020}, dhSegment \cite{dhsegment} and ARU-Net \cite{gruning2018}. These models need pixel-level annotated images for training, we present in the next paragraphs the challenges faced to generate a unified annotated training set.

\subsubsection{Diversity in the annotations}

To generate the pixel-level label images, the bounding polygons are extracted from the ground-truth and drawn on a black background image of the same size. As one can see on Figure \ref{fig:datasets}, the annotations are really diverse among the datasets:
\begin{itemize}
    \item In most datasets (AN-Index, Balsac, Bozen, BNPP, HOME and Horae), the images are annotated using simple polygons including line ascenders and descenders;
    \item In cBAD2019, READ-Simple and READ-Complex, the ascenders and descenders are mostly not included in the polygons that are really thin compared to the first case;
    \item In ScribbleLens images, the annotations are wide rectangles that can contain a lot of background;
    \item In DIVA-HisDB text lines are annotated using more complex polygons following carefully each letter contour as well as Alcar lines but with coarser contours.
\end{itemize}

This diversity in the annotations prevents us from directly training a generic model that could be applied to new datasets, as the annotation would sometimes be inconsistent between two examples from two different datasets. This would dramatically degrade the performances of a system trained with such inconsistencies. Correcting the label inconsistencies between the datasets is a necessity to allow the unification of the training sets.

\subsubsection{Overlapping polygons}

\begin{figure}
    \begin{center}
        \includegraphics[width=0.4\textwidth]{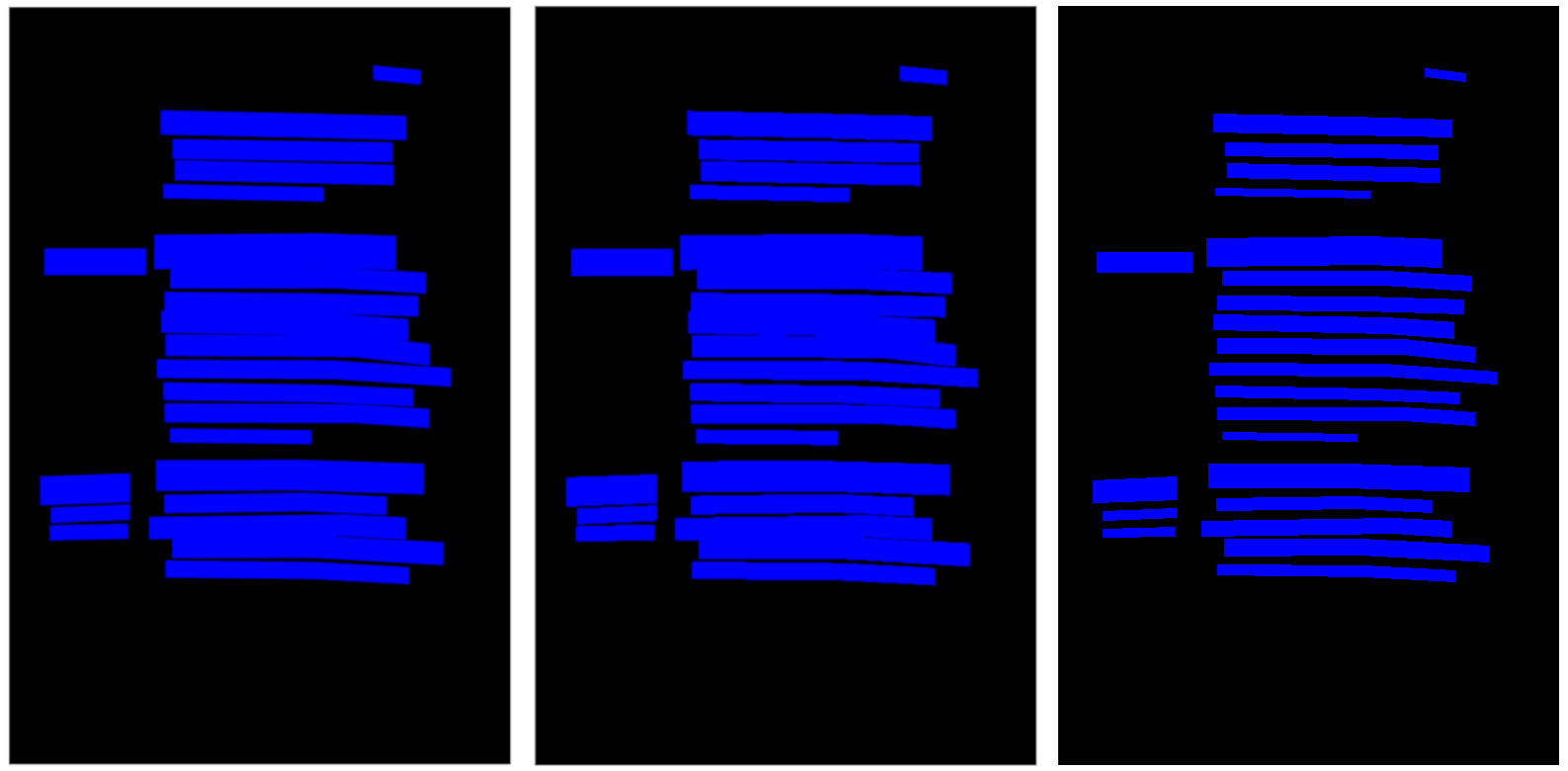}
    \end{center}
    \caption{Label generation process for an image from the Bozen dataset. Left: label image generation at the original image size. Center: rescaling to the network input size. Right: rescaling the bounding polygons at the network input size, then split and label image generation at the network input size.}
    \label{fig:bad_label}
\end{figure}

\sloppy
Another challenge accentuating the label inconsistencies is the overlap of annotated bounding polygons. Even if some datasets have been annotated such that the polygons never touch nor overlap, some others have been less carefully annotated, leading to touching and overlapping polygons (ScribbleLens page on Figure \ref{fig:datasets}). When drawn on an image, the corresponding polygons also overlap, as shown on the left image of Figure \ref{fig:bad_label}. Obviously, such a ground-truth cannot serve for a precise evaluation of the capacity of a system to detect each line of text.
Moreover, both Doc-UFCN and ARU-Net take sub-resolutions of the input images, which may lead to some additional mergers in the annotation images. \\ Figure \ref{fig:bad_label} presents this undesirable effect by showing the original annotation image (left) and its resized version at \textit{768} pixels (center).
\fussy
In the literature, most studies use the ground-truth as is without paying much attention to this merger problem. The main reason is probably because the evaluation metrics used only count pixel accuracy, without evaluating the line detection process more precisely. However, a significant number of mergers in the training dataset will definitely skew the network towards merging more lines than desired, with a negative induced effect on the HTR system which is unable to recognize vertically merged lines.

For our experiments, we mitigated this problem by splitting as much as possible the overlapping lines, without losing too much information.

\subsection{Unifying labeling strategy}

To better unify the annotations, we chose to have only simple polygons during training. Therefore, the enclosing bounding rectangles of DIVA-HisDB polygons have been used during training instead of the original complex polygons. In addition, to solve the problem of overlapping bounding polygons of some datasets, we used the following strategy. For a given image, we look for the touching and overlapping lines pairs. Then three cases have been identified for each pair:
\begin{itemize}
    \item The polygons are touching: we erode them by \textit{1} px;
    \item The polygons are overlapping by less than \textit{20}\% each: we apply the splitting process described below;
    \item The polygons are overlapping by more than \textit{20}\%: we keep them as such since the splitting may lead to loosing too much information (loss of a polygon or separation into two polygons).
\end{itemize}

In the case of a small overlap (less than \textit{20}\%) the following process is applied: we look for the line with the smallest area of overlap and remove its intersection with the other line, while the other line is kept as is. In this end, all these polygons are drawn over a black background image. This threshold of \textit{20}\% has been chosen since it is about the height of the ascenders and descenders in the polygons.

Since resizing can cause undesirable mergers, the bounding polygons are first rescaled to the size of the input image of the network, and the split is applied at this scale. That way, the annotation image is directly generated at the desired resolution input image of the network, which prevents from re-merging some ground-truth lines. The right image of Figure \ref{fig:bad_label} shows the result of this process. As one can see, the label image produced contains polygons which are better separated. Even if some overlapping lines may still remain on some pages, we expect to have generated a more suitable ground-truth that will help training the segmentation model and improve its ability to predict separated text lines. The code to generate these modified annotations and the label images used in the experiments are publicly available\footnote{\url{https://gitlab.com/teklia/dla/arkindex_document_layout_training_label_normalization}}.

\section{Benchmarking segmentation approaches}
\label{sec:training_details}
For our experiments, we chose to study three state-of-the-art systems: Doc-UFCN \cite{boillet2020}, dhSegment \cite{dhsegment} and ARU-Net \cite{gruning2018}. Doc-UFCN has been chosen for its good performances on historical datasets \cite{boillet2020} while using fewer parameters and requiring less computing for the inference. In addition, we selected dhSegment and ARU-Net since they are open-source, the systems are easy to train, and they have shown good performance on historical document segmentation tasks. ARU-Net is also the line segmentation model used in Transkribus \cite{transkribus}, the most popular platform for historical document processing. We now present the systems and the training details, since we trained them all so as to fairly compare their performance.

\subsection{Doc-UFCN}

Doc-UFCN is a U-shaped Fully Convolutional Network. It contains an encoder, whose goal is to represent the input image by \textit{256} feature maps, that consists in dilated convolutions followed by max pooling layers. The decoder's aim is to reconstruct the input image with a pixel-wise labeling. It contains blocks of one standard convolution followed by one transposed convolution. Finally, the features computed during the encoding step are concatenated with those of the decoding stage.

To reduce the training time while keeping relevant information, we chose to train Doc-UFCN on resized input images such that, according to experiments in \cite{boillet2020}, their largest dimension is \textit{768} pixels while preserving their original aspect ratio. Therefore, to train Doc-UFCN, the annotations are directly generated at \textit{768} pixels using the process presented above.
For the following experiments, Doc-UFCN is trained with an initial learning rate of \textit{5e-3}, mini-batches of size \textit{2}, Adam optimizer, Dice loss and early stopping. According to \cite{boillet2020}, all dropout layers have the same probability of \textit{0.4}. The code of this system is now publicly available\footnote{\url{https://pypi.org/project/doc-ufcn/}}.

\begin{table}
    \caption{Comparison of the three systems: number of parameters and mean inference time in second/image measured on the Balsac dataset. The predictions have been made using a GPU GeForce RTX 2070 8G. dhSegment has 32.8M parameters, but since the encoder is pre-trained only 9.36M parameters have to be fully-trained.}
    \setlength\tabcolsep{3.5pt}
    \label{tab:sys_comp}
    \begin{center}
        \begin{tabular}{l|rrr}
            \hline\noalign{\smallskip}
            \multicolumn{1}{c}{} & Doc-UFCN & dhSegment & ARU-Net \\ \noalign{\smallskip}\hline\noalign{\smallskip}
            Nb of parameters & \color{best_result}\underline{4.09M} & 32.8M (9.36M) & 4.14M \\
            Inference time & \color{best_result}\underline{0.41} & 2.95 & 1.39 \\ \noalign{\smallskip}
            \hline
        \end{tabular}
    \end{center}
\end{table}

\subsection{dhSegment}

dhSegment's architecture is deeper than Doc-UFCN one. The encoder is composed of convolution and pooling layers (up to \textit{2048} feature maps) and has been pre-trained on natural scene images \cite{ImageNet}. The decoder is similar to the previous one by using successive blocks of one standard convolution and one upscaling layer.

By contrast with the two other systems, dhSegment is trained on patches of full-size images, and therefore the splitting process is applied on original size polygons. The model is trained with early stopping, without any resizing of the images but using patches of \textit{400$\times$400} pixels and mini-batches of size \textit{4}. In addition, we kept the post-processing proposed in \cite{dhsegment} by thresholding the probabilities with \textit{t = 0.7}. Different values have been tested for this parameter, \textit{0.7} gave the best results on the validation set.

\subsection{ARU-Net}

ARU-Net is an extended version of the standard U-Net \cite{unet}. Two concepts have been added: a spatial attention and a residual structure. The spatial attention (A-Net) is a multi-layer CNN and is used to handle various font sizes on a single page. The residual blocks are introduced to enable the error  backpropagation and identity propagation.

For ARU-Net annotations, we use the same process as for Doc-UFCN, but on polygons at \textit{33}\% of their original size. ARU-Net is trained with the rescaled images using early stopping. We used the Cross Entropy Loss and an initial learning rate of \textit{1e-3}. As for dhSegment, one should threshold the probabilities to have the final predictions. However, choosing this parameter for ARU-Net was not an easy task since it really impacts the results. In the end, we chose a low threshold of \textit{t = 0.3} since a higher value would have removed a substantial amount of text line pixels. \\

For all the architectures, we keep the parameters of the models with the lowest validation loss as those of the best model. In addition, small connected components are detected on the predictions and those smaller than \textit{min\_cc = 50} pixels are removed. This value also gave the best results on the validation set.

\subsection{Models comparison}

Table \ref{tab:sys_comp} shows the number of parameters and inference times of the three systems. Doc-UFCN and ARU-Net have similar weights in number of parameters, while being way lighter than dhSegment. For the inference times, dhSegment and ARU-Net are off-track, being really slower than Doc-UFCN.

\begin{table*}
    \caption{Results obtained at pixel-level on the testing sets with the labels unification process by Doc-UFCN (UFCN), dhSegment and ARU-Net. UFCN \xmark{} shows the results of Doc-UFCN when trained on original non-normalized labels. ScribbleLens* reports the results of specific models.}
    \label{tab:res_iou}
    \begin{center}
        \begin{tabular}{l|rrrr|rrrr}
            \hline\noalign{\smallskip}
            \multicolumn{1}{c}{\multirow{2}{*}{\textsc{Dataset}}} & \multicolumn{4}{c}{\textsc{IoU}} & \multicolumn{4}{c}{\textsc{F1-score}} \\
            \multicolumn{1}{c}{} & \multicolumn{1}{c}{UFCN} & dhSegment & ARU-Net & \multicolumn{1}{c}{UFCN \xmark{}} & UFCN & dhSegment & ARU-Net & \multicolumn{1}{c}{UFCN \xmark{}} \\ \noalign{\smallskip}\hline\noalign{\smallskip}
            AN-Index & \color{best_result}\underline{0.69} & 0.68 & 0.68 & 0.67 & \color{best_result}\underline{0.82} & 0.81 & 0.74 & 0.80 \\
            Balsac & \color{best_result}\underline{0.87} & 0.74 & 0.84 & 0.71 & 0.93 & 0.85 & \color{best_result}\underline{0.98} & 0.83 \\
            BNPP & 0.65 & 0.60 & \color{best_result}\underline{0.75} & 0.63 & 0.78 & 0.75 & \color{best_result}\underline{0.90} & 0.77 \\
            Bozen & \color{best_result}\underline{0.82} & 0.70 & 0.74 & 0.67 & 0.90 & 0.82 & \color{best_result}\underline{0.99} & 0.80 \\
            cBAD2019 & 0.66 & 0.62 & \color{best_result}\underline{0.74} & 0.61 & 0.79 & 0.76 & \color{best_result}\underline{0.89} & 0.75 \\
            DIVA-HisDB & \color{best_result}\underline{0.67} & 0.46 & 0.60 & 0.65 & 0.80 & 0.60 & \color{best_result}\underline{0.96} & 0.78 \\
            HOME & 0.60 & 0.55 & \color{best_result}\underline{0.67} & 0.56 & 0.77 & 0.73 & \color{best_result}\underline{0.94} & 0.74 \\
            Horae & 0.64 & 0.63 & \color{best_result}\underline{0.75} & 0.64 & 0.79 & 0.79 & \color{best_result}\underline{0.87} & 0.79 \\
            READ-Complex & 0.49 & 0.58 & \color{best_result}\underline{0.73} & 0.53 & 0.70 & 0.73 & \color{best_result}\underline{0.81} & 0.74 \\
            READ-Simple & 0.60 & 0.57 & \color{best_result}\underline{0.71} & 0.58 & 0.73 & 0.71 & \color{best_result}\underline{0.88} & 0.72 \\
            \textbf{mean} & \textbf{0.67} & \textbf{0.61} & \textbf{\color{bf_best_result}\underline{0.72}} & \textbf{0.63} & \textbf{0.80} & \textbf{0.76} & \textbf{\color{bf_best_result}\underline{0.90}} & \textbf{0.78} \\
            \noalign{\smallskip}\hline\noalign{\smallskip}
            ScribbleLens & 0.35 & 0.36 & 0.41 & \color{best_result}\underline{0.42} & 0.51 & 0.51 & 0.58 & \color{best_result}\underline{0.59} \\
            Alcar & 0.35 & 0.49 & \color{best_result}\underline{0.58} & 0.51 & 0.49 & 0.60 & \color{best_result}\underline{0.70} & 0.63 \\ \noalign{\smallskip}\hline\noalign{\smallskip}
            ScribbleLens* & 0.80 & \color{best_result}\underline{0.95} & -~~ & -~~ & 0.89 & \color{best_result}\underline{0.97} & -~~ & -~~ \\ \noalign{\smallskip}
            \hline
        \end{tabular}
    \end{center}
\end{table*}

\section{Segmentation evaluation}
\label{sec:seg_eval}
\sloppy
The three models have been trained on all the training images to have generic models. In this section, we report the results at pixel and object-levels. To have a fair comparison, all the predictions are first resized to the original image size before running the evaluation. In addition, to be comparable to other results published in the literature, the models are evaluated using the original ground-truth lines. It is worth noting that this evaluation based on original labels will definitively decrease the evaluation values since the training polygons are way thinner than the ground-truth ones, while allowing sound visual results. Since we aimed at developing a generic historical model showing good performances on unseen datasets, we also report the results on ScribbleLens and Alcar, two datasets that were not used for training. Lastly, we show the impact of the labels unification on the segmentation results when training Doc-UFCN system.

\subsection{Pixel-level metrics}
\label{sec:pixel_res}

\fussy
As shown in Table \ref{tab:metrics}, most systems of the literature are evaluated using pixel-level metrics. Precision (P) and Recall (R) are first computed using equations \ref{precision_recall}, then the Intersection-over-Union (IoU) as well as the F1-score are computed using equations \ref{iou} and \ref{f1}.

\vspace{-0.45cm}
\begin{equation}
    \hspace{1cm} \text{P} = \frac{TP}{TP+FP}   \hspace{1.2cm}    \text{R} = \frac{TP}{TP+FN}
    \label{precision_recall}
\end{equation}
\vspace{-0.45cm}
\begin{equation}
    \hspace{1cm} \text{IoU} = \frac{TP}{TP+FP+FN}
    \label{iou}
\end{equation}
\vspace{-0.45cm}
\begin{equation}
    \begin{split}
        \hspace{1cm} \text{F1-score} = \frac{2 \times TP}{2 \times TP+FP+FN} \\
        & \hspace*{-3.68cm} = 2 \times \frac{P \times R}{P + R}
    \end{split}
    \label{f1}
\end{equation}

These equations apply at pixel-level with:
\vspace{-0.2cm}
\begin{itemize}
    \item \textit{TP}: number of positive pixels correctly predicted;
    \item \textit{FP}: number of negative pixels predicted as positive; 
    \item \textit{FN}: number of positive pixels predicted as negative.
\end{itemize}

\subsubsection{Systems comparison on the training datasets}

\begin{figure}
    \vspace{9cm}
    \centering
        \def\svgwidth{0.42\textwidth}
        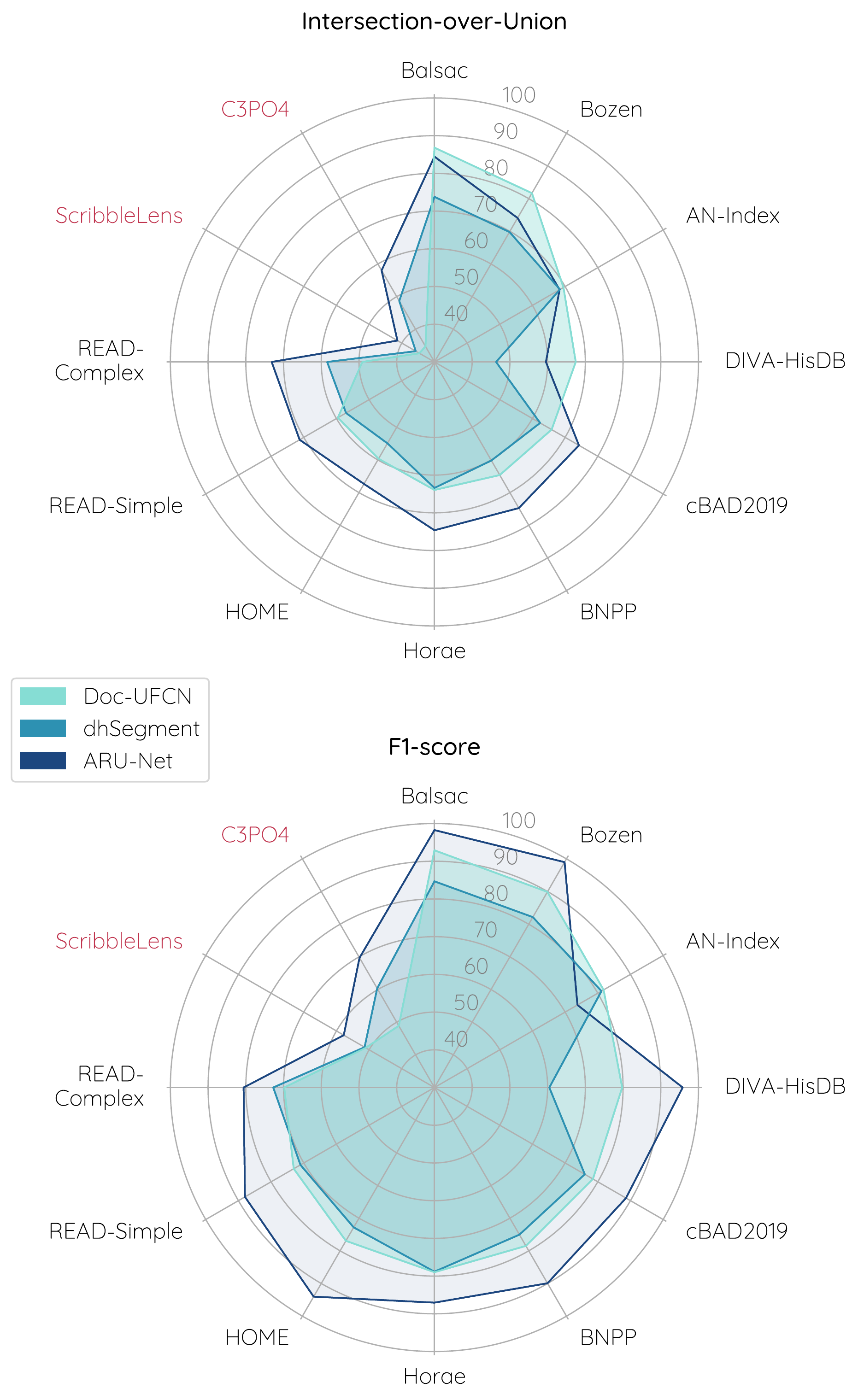 
    \caption{Results of the three networks at pixel-level on the test sets (\%).}
    \label{fig:res_iou}
\end{figure}

The results obtained by the three networks on the testing sets are shown at the top of Table \ref{tab:res_iou} and summarized on Figure \ref{fig:res_iou}. ARU-Net seems to perform better in terms of F1-score, whereas it is more mixed considering the IoU. Figure \ref{fig:res_iou} also highlights the performance shift of ARU-Net between its IoU and F1 scores. Indeed, the F1-score really relies on the Precision and Recall measures (not presented here), accurately summarizing them. This can explain the high results ARU-Net obtains, since its Recall values are high for all the datasets. On the contrary, the IoU score being less focused on the correctly predicted pixels (\textit{TP} is considered twice for F1-score and only once for IoU), IoU scores are lower, which leads to a lower ranking in the table of results.

\begin{figure*}
    \centering
        \def\svgwidth{0.85\textwidth}
        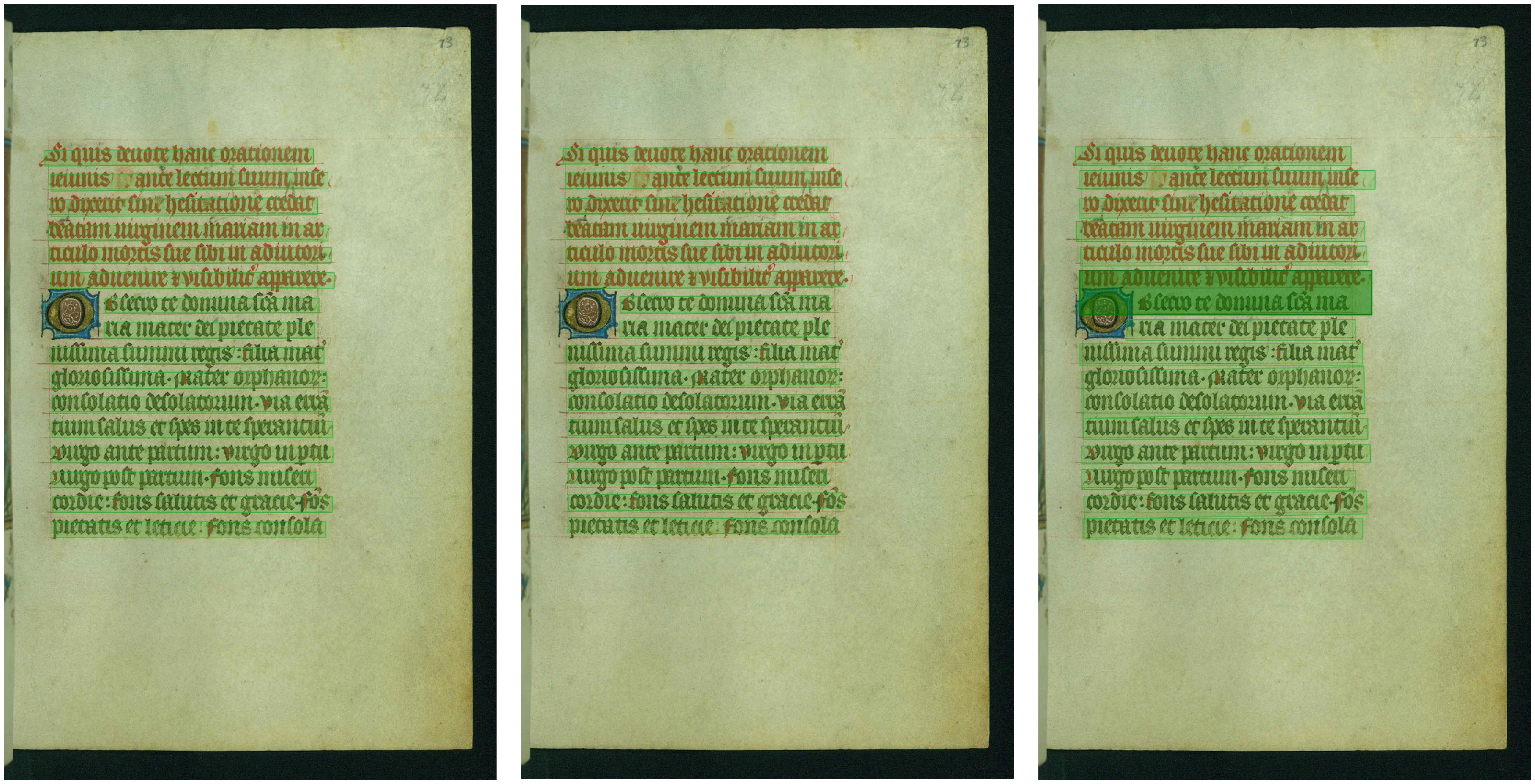
    \caption{Predictions of a page from Horae dataset: Doc-UFCN on the left, dhSegment in the middle and ARU-Net on the right. Doc-UFCN and dhSegment output similar results while ARU-Net shows thicker lines and two mergers of two lines (one is highlighted is dark green).}
    \label{fig:res_horae_visualization}
\end{figure*}

These low Precision but high Recall values obtained by ARU-Net suggest that the model has correctly retrieved the majority of text line pixels, while at the same time many background pixels have been classified as text line. This reflects the presence of mergers in the detected lines. Figure \ref{fig:res_horae_visualization} shows the predictions obtained by the networks on a randomly chosen image from the Horae test set. It confirms our hypothesis that ARU-Net has merged some lines. However, as stated before, using a higher threshold would have led to miss a lot of text line pixels. We believe that ARU-Net is maybe not the most appropriate system for detecting close objects. Indeed, it has often shown really good performances when trained using baseline annotations where the objects are more spaced and thin than text line bounding polygons.

Comparing Doc-UFCN and dhSegment is a bit easier since they behave alike for the IoU and F1 scores. Doc-UFCN outperforms dhSegment on the majority of the datasets for both measures. It is however the worst model on the READ-Complex dataset. We assume that it is due to the high number of very small objects in the document images that can have been missed by Doc-UFCN since it works at a really low resolution by contrast to dhSegment and ARU-Net. \\

Evaluating and comparing the three models based on the IoU and F1-score only would lead to chose ARU-Net as the best model, whereas we have shown that its low precisions can lead to a really low capacity to distinguish close lines. Object-level metrics, that can account for the merged lines, should be used as a complement to these pixel values.

\subsubsection{Generalization to unseen datasets}

Table \ref{tab:res_iou} also shows the results of the generic models applied to ScribbleLens and Alcar datasets. We also trained Doc-UFCN and dhSegment on ScribbleLens to have specific models for comparison. For the Alcar data-set, we do not have any line segmentation training images, therefore only the generic results are presented.

The values obtained by the three systems on ScribbleLens and Alcar datasets are way lower than those obtained on the training datasets and also those obtained by the specific ScribbleLens* models. For the ScribbleLens test set, the precision is equal to \textit{97\%} for dhSegment and Doc-UFCN generic models, while it is between \textit{82} and \textit{85\%} for Alcar. This suggests that almost all the predicted pixels were correct, whereas a lot of ground-truth pixels were missed. Our hypothesis is that the models predicted good but really thin polygons compared to really wide annotation polygons of ScribbleLens pages, leading to degraded IoU values. The same applies for Alcar images, where thin rectangle-like polygons including only few background pixels have probably been predicted.

\sloppy
Based on these metrics, we cannot be certain whether the systems fail to generalize on the two new datasets. Additional metrics might give more insight on the actual generalization capacities of the models.

\begin{figure*}
    \begin{center}
        \subfloat[Label image.]{
            \includegraphics[width=0.26\textwidth]{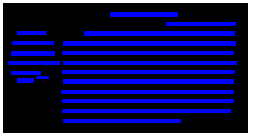}
            \label{fig:bad_metric_label}
        }
        \vspace{-0.65cm}
        \hspace{0.4cm}
        \includegraphics[width=0.26\textwidth]{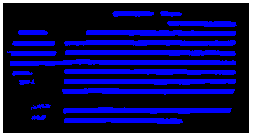}
        \hspace{0.4cm}
        \includegraphics[width=0.26\textwidth]{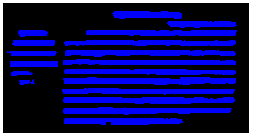}
        \hspace*{0.3\textwidth}
        \subfloat[First prediction: \\
            \hspace*{0.55cm} IoU = 0.72 \hspace{0.5cm} F1 = 0.84 \\
            \hspace*{0.55cm} P \hspace{0.195cm} = 0.81 \hspace{0.5cm} R \hspace{0.028cm} = 0.87 \\
            \hspace*{0.55cm} AP@.5 = \textbf{0.68}]{
            \includegraphics[width=0.26\textwidth]{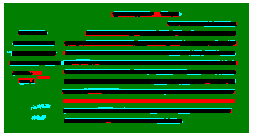}
            \label{fig:bad_metric_pred1}
        }
        \hspace{0.3cm}
        \subfloat[Second prediction: \\
            \hspace*{0.55cm} IoU = 0.72 \hspace{0.5cm} F1 = 0.84 \\
            \hspace*{0.55cm} P \hspace{0.195cm} = 0.75 \hspace{0.5cm} R \hspace{0.028cm} = 0.95 \\
            \hspace*{0.55cm} AP@.5 = \textbf{0.94}]{
            \includegraphics[width=0.26\textwidth]{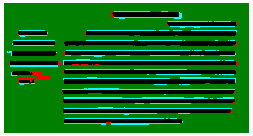}
            \label{fig:bad_metric_pred2}
        }
    \end{center}
    \caption{Two different predictions for the same image showing the same IoU and F1 measures. Overlaps are generated with DIVA tool \cite{albertiDIVA}. Green and black respectively correspond to correctly predicted background and foreground pixels. Cyan represents false positive pixels and red false negative ones. Here, only the object-level AP score (with an IoU threshold of \textit{50}\%) allows to accurately evaluate and compare the predictions.}
    \label{fig:bad_metric}
\end{figure*}

\subsubsection{Impact of the annotations unification}

To evaluate the impact of the labels unification on the results, we trained Doc-UFCN on all the datasets with non-normalized annotations. According to Table \ref{tab:res_iou}, training with the normalized ground-truth improves the performances at pixel-level by up to +16 percentage points of IoU on Balsac. However, as explained for ARU-Net in Section \ref{sec:pixel_res}, the Recall measures are higher without the unification process. Indeed, the pixels between consecutive lines and the ones along line edges are more often predicted as text lines, increasing the Recall values. However, some of these pixels are not supposed to be part of the text lines (since they create some mergers), which decreases the Precision values. Based on these metrics, splitting close lines appears to be necessary to help the model distinguishing nearby lines.

\subsubsection{Limitations of the pixel-level metrics}

\fussy
Even if these pixel-level measures can give a first idea on how a model performs, we show on Figure \ref{fig:bad_metric} that they might not be sufficient. Indeed, Figures \ref{fig:bad_metric_pred1} and \ref{fig:bad_metric_pred2} present two predictions for the same image (at the top) and their overlap (at the bottom) with the ground-truth label shown on Figure \ref{fig:bad_metric_label}. The first one shows split and merged lines, a missing line (in red) and some false positives (in cyan). On the contrary, the second one shows thicker lines but no missing nor false positives. In that way, the second prediction seems better. However, the IoU and F1-score are equals for both predictions. In the literature, systems are often compared using IoU and F1-score values, we show here that these metrics are not suitable for choosing the best model since they do not consider the number of detected objects. In conclusion, these metrics do not allow us to determine the generalization capacity of the trained models. To overcome these problems, the next Section \ref{sec:object_level} presents and analyzes the results using object-level metrics. 

\begin{table*}
    \setlength\tabcolsep{4.5pt}
    \caption{Results obtained at line-level on the testing sets with the labels unification process by Doc-UFCN (UFCN), dhSegment (dhSeg) and ARU-Net. UFCN \xmark{} shows the results of Doc-UFCN when trained on original non-normalized labels. ScribbleLens* reports the results of specific models.}
    \label{tab:res_obj}
    \begin{center}
        \begin{tabular}{l|rrrr|rrrr|rrrr}
            \hline\noalign{\smallskip}
            \multicolumn{1}{c}{\multirow{2}{*}{\textsc{Dataset}}} & \multicolumn{4}{c}{\textsc{AP@.5}} & \multicolumn{4}{c}{\textsc{AP@.75}} & \multicolumn{4}{c}{\textsc{AP@[.5, .95]}} \\
            \multicolumn{1}{c}{} & \multicolumn{1}{c}{UFCN} & dhSeg & ARU & \multicolumn{1}{c}{UFCN \xmark{}} & UFCN & dhSeg & ARU & \multicolumn{1}{c}{UFCN \xmark{}} & UFCN & dhSeg & ARU & \multicolumn{1}{c}{UFCN \xmark{}} \\ \noalign{\smallskip}\hline\noalign{\smallskip}
            AN-Index & 0.75 & \color{best_result}\underline{0.76} & 0.51 & 0.69 & \color{best_result}\underline{0.27} & 0.23 & 0.03 & 0.17 & 0.34 & \color{best_result}\underline{0.35} & 0.17 & 0.28 \\
            Balsac & \color{best_result}\underline{0.98} & 0.94 & 0.76 & 0.95 & \color{best_result}\underline{0.91} & 0.54 & 0.20 & 0.24 & \color{best_result}\underline{0.76} & 0.51 & 0.34 & 0.44 \\
            BNPP & \color{best_result}\underline{0.83} & 0.78 & 0.50& 0.81 & \color{best_result}\underline{0.12} & 0.11 & 0.02 & \color{best_result}\underline{0.12} & \color{best_result}\underline{0.31} & 0.27 & 0.13 & 0.30 \\
            Bozen & \color{best_result}\underline{0.99} & 0.74 & 0.01 & 0.77 & \color{best_result}\underline{0.85} & 0.22 & 0.0 & 0.11 & \color{best_result}\underline{0.69} & 0.35 & 0.0 & 0.31 \\
            cBAD2019 & \color{best_result}\underline{0.86} & 0.71 & 0.29 & 0.71 & \color{best_result}\underline{0.50} & 0.11 & 0.01 & 0.14 & \color{best_result}\underline{0.48} & 0.24 & 0.07 & 0.25 \\
            DIVA-HisDB & 0.77 & 0.39 & 0.10 & \color{best_result}\underline{0.86} & \color{best_result}\underline{0.33} & 0.11 & 0.03 & 0.27 & 0.36 & 0.17 & 0.04 & \color{best_result}\underline{0.40} \\
            HOME & \color{best_result}\underline{0.85} & 0.78 & 0.19 & 0.82 & \color{best_result}\underline{0.49} & 0.12 & 0.00 & 0.18 & \color{best_result}\underline{0.46} & 0.28 & 0.04 & 0.33 \\
            Horae & 0.83 & \color{best_result}\underline{0.85} & 0.56 & 0.84 & \color{best_result}\underline{0.31} & 0.17 & 0.08 & 0.18 & \color{best_result}\underline{0.38} & 0.34 & 0.17 & 0.34 \\
            READ-Complex & 0.60 & \color{best_result}\underline{0.62} & 0.22 & 0.61 & 0.11 & 0.15 & 0.08 & \color{best_result}\underline{0.16} & 0.23 & \color{best_result}\underline{0.24} & 0.08 & \color{best_result}\underline{0.24} \\
            READ-Simple & \color{best_result}\underline{0.69} & 0.58 & 0.21 & 0.60 & \color{best_result}\underline{0.17} & 0.12 & 0.01 & 0.09 & \color{best_result}\underline{0.28} & 0.21 & 0.05 & 0.19 \\
            \textbf{mean} & \textbf{\color{bf_best_result}\underline{0.82}} & \textbf{0.72} & \textbf{0.34} & \textbf{0.77} & \textbf{\color{bf_best_result}\underline{0.41}} & \textbf{0.19} & \textbf{0.05} & \textbf{0.17} & \textbf{\color{bf_best_result}\underline{0.43}} & \textbf{0.30} & \textbf{0.11} & \textbf{0.31} \\
            \noalign{\smallskip}
            \hline\noalign{\smallskip}
            ScribbleLens & 0.06 & 0.02 & 0.0 & \color{best_result}\underline{0.41} & 0.0 & 0.0 & 0.0 & 0.0 & 0.02 & 0.02 & 0.0 & \color{best_result}\underline{0.08} \\
            Alcar & 0.16 & 0.76 & 0.0 & \color{best_result}\underline{0.86} & 0.00 & \color{best_result}\underline{0.10} & 0.0 & 0.06 & 0.03& 0.26 & 0.0 & \color{best_result}\underline{0.27} \\ \noalign{\smallskip}
            \hline\noalign{\smallskip}
            ScribbleLens* & \color{best_result}\underline{0.94} & 0.0 & -~~ & -~~ & \color{best_result}\underline{0.82} & 0.0 & -~~ & -~~ & \color{best_result}0.61 & 0.0 & -~~ & -~~ \\ \noalign{\smallskip}
            \hline
        \end{tabular}
    \end{center}
\end{table*}

\subsection{Object-level metrics}
\label{sec:object_level}

We showed in the previous section that the pixel-level metrics may not be sufficient for in depth evaluation and comparison of models. We now present object-level metrics and show that they are complementary to the previous metrics and can give more useful information about the quality of a segmentation result. \\

As stated before, determining if an object should be considered positive or negative is not simple. Based on the idea proposed in the PASCAL VOC Challenges, one can compute the Precision, Recall, and Average Precision (AP) at object-level. To do so, the predicted and ground-truth objects are first paired according to their IoU scores such that only one predicted object can be paired with a ground-truth one and conversely.

Then, the paired objects are ranked by decreasing confidence score, and Precision P$_{k}$ and Recall R$_{k}$ measures are computed for each confidence rank \textit{k}, depending on a chosen IoU threshold \textit{t}, using equations \ref{rank_precision_recall}.

\vspace{-0.5cm}
\begin{equation}
    \hspace{1cm} \text{P}_k = \frac{TP_k}{Total_k}   \hspace{1.7cm}    \text{R}_k = \frac{TP_k}{Total_{GT}} \\
    \label{rank_precision_recall}
\end{equation}

These equations apply with:
\vspace{-0.1cm}
\begin{itemize}
    \item \textit{TP$_{k}$}: number of positive objects correctly predicted above rank \textit{k};
    \item \textit{Total$_{k}$}: number of predicted objects above rank \textit{k}; 
    \item \textit{Total$_{GT}$}: number of ground-truth objects to retrieve;
\end{itemize}
\vspace{-0.1cm}
where an object is considered positive if its IoU is higher than the chosen threshold \textit{t}.

The Precision-Recall curve is then computed and interpolated, and the Average Precision (AP) is defined as the area under this curve. This AP is computed for all classes of an experiment and then averaged over the classes, giving the mean Average Precision (mAP). For text line detection, we only have one object class therefore the mAP is equal to the AP and is denoted as AP@\textit{t} in the following, \textit{t} still being the IoU threshold.

\subsubsection{Systems comparison on the training datasets}

\begin{figure*}
    \begin{center}
        \def\svgwidth{0.95\textwidth}
        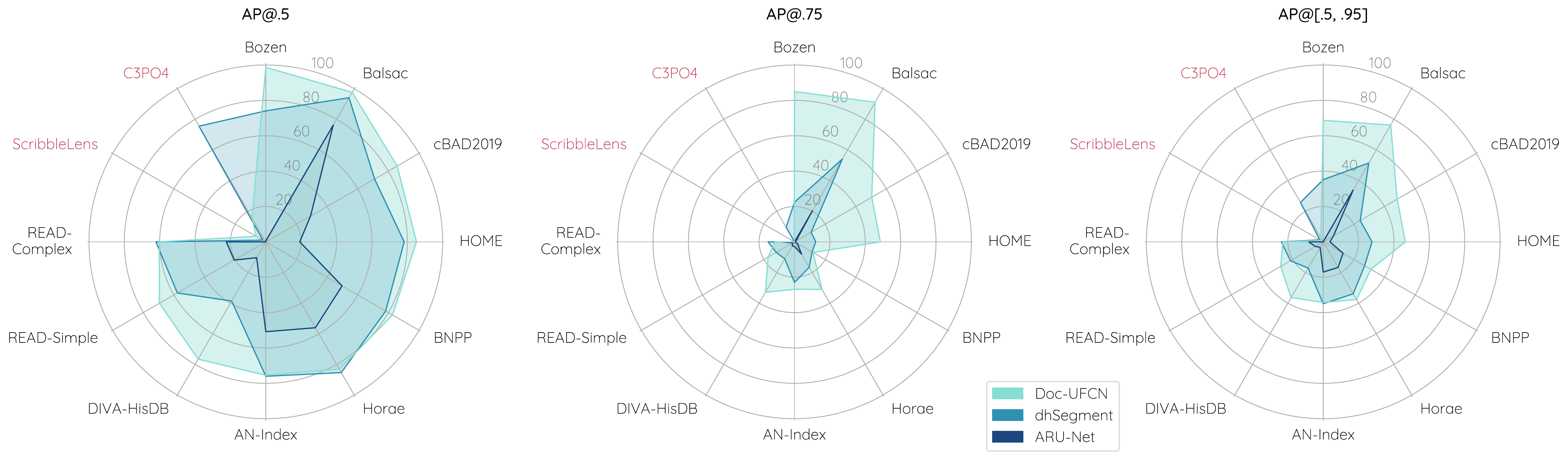 
    \end{center}
    \caption{Results of the three networks at line-level on the test sets (\%).}
    \label{fig:res_aps}
\end{figure*}

Table \ref{tab:res_obj} and Figure \ref{fig:res_aps} present the AP results obtained on the testing sets for two IoU thresholds: \textit{50}\% (AP@.5) and \textit{75}\% (AP@.75). In addition and to get rid of any threshold, AP averaged over a range of IoU values (\textit{50}\%-\textit{95\%}) is also computed and presented as AP@[.5, .95]. 

The results presented here reinforce our previous hypothesis that ARU-Net fails to split close objects. Indeed, all the ARU-Net results are much lower compared to those of the two other systems, except for the Balsac dataset where the text line polygons are really spaced in the annotations.

\begin{figure*}
    \centering
        \def\svgwidth{0.7\textwidth}
        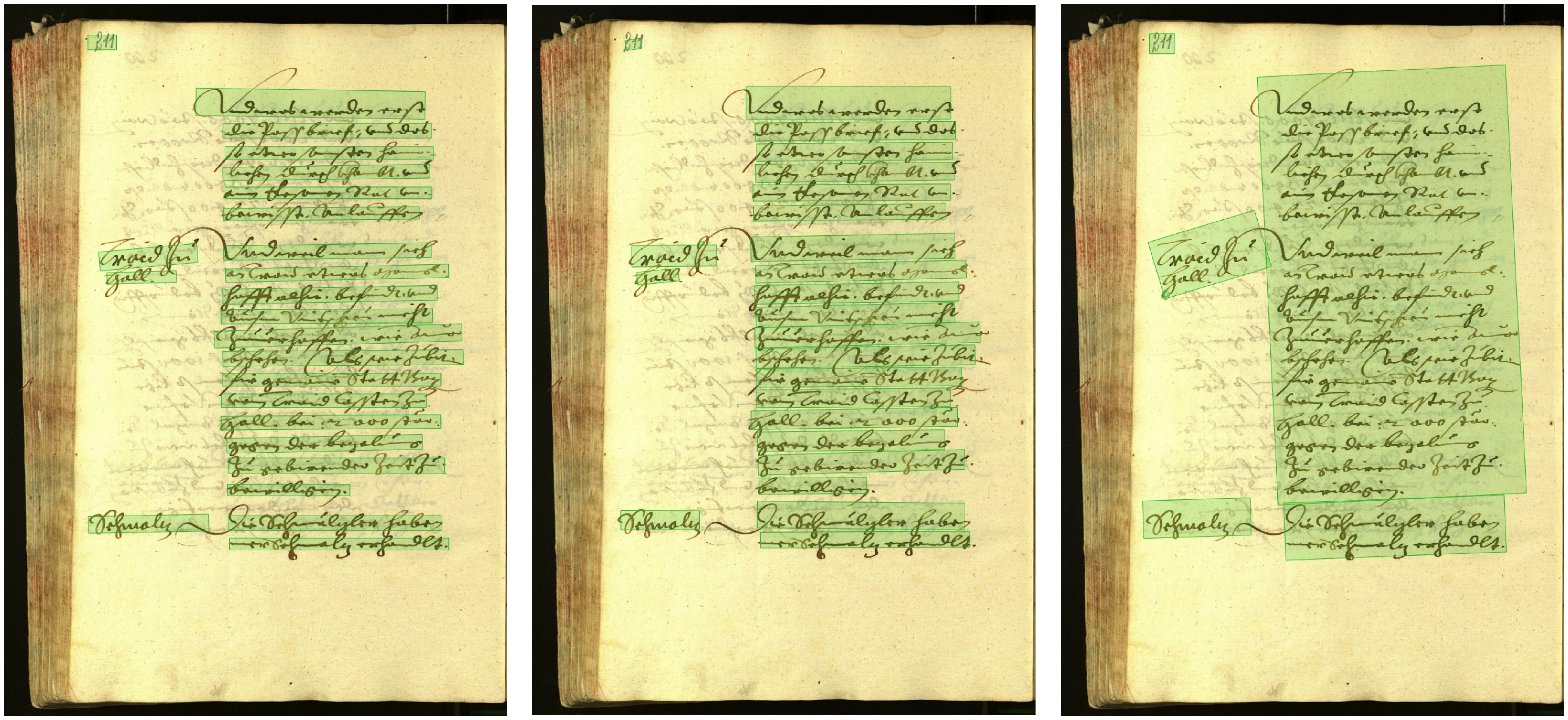
    \caption{Predictions of a page from Bozen dataset: Doc-UFCN on the left, dhSegment in the middle and ARU-Net on the right. Doc-UFCN shows well split lines, while ARU-Net still shows merged lines. dhSegment does not output any merged lines; however, they are closer than the ones output by Doc-UFCN. In addition, dhSegment polygons include more space at the top of the lines, which might have a negative impact on text recognition.}
    \label{fig:res_bozen_visualization}
\end{figure*}

In addition, we see on Figure \ref{fig:res_aps} that for a low threshold of \textit{50}\%, Doc-UFCN slightly outperforms dhSegment. When moving from \textit{50}\% to \textit{75}\%, one can see that the results of both models are degrading, meaning that some lines are now considered as false positives because their localization is not accurate enough (less than \textit{75}\% of IoU). However, this degradation is lower for Doc-UFCN than for dhSegment, meaning that the localization of dhSegment objects is less accurate than that of Doc-UFCN. The same conclusion applies when looking at the AP[.5, .95], meaning that Doc-UFCN is more precise (less missed and false positive objects) and accurate (better localization) than dhSegment. Figure \ref{fig:res_bozen_visualization} shows the results of the three networks on a randomly chosen image from the Bozen dataset. These predictions confirm the interest of AP measures to correctly evaluate segmentation predictions, since it highlights bad behaviors like those shown by ARU-Net.

\subsubsection{Generalization to unseen datasets}

Table \ref{tab:res_obj} also shows the results of the generic models applied to ScribbleLens and Alcar and the specific ScribbleLens models. As contrast with previous pixel-level table, the results obtained by the specific models are totally opposing. The results obtained by Doc-UFCN confirmed that the model is working well when trained directly on ScribbleLens. On the contrary, whereas dhSegment showed good pixel-level measures, its object values are all at \textit{0}. Indeed, as for the previous results obtained by ARU-Net, there are a lot of merged lines predicted by the dhSegment specific model, meaning that it failed to learn directly from ScribbleLens images.

\begin{figure}
    \begin{center}
        \includegraphics[width=0.2\textwidth]{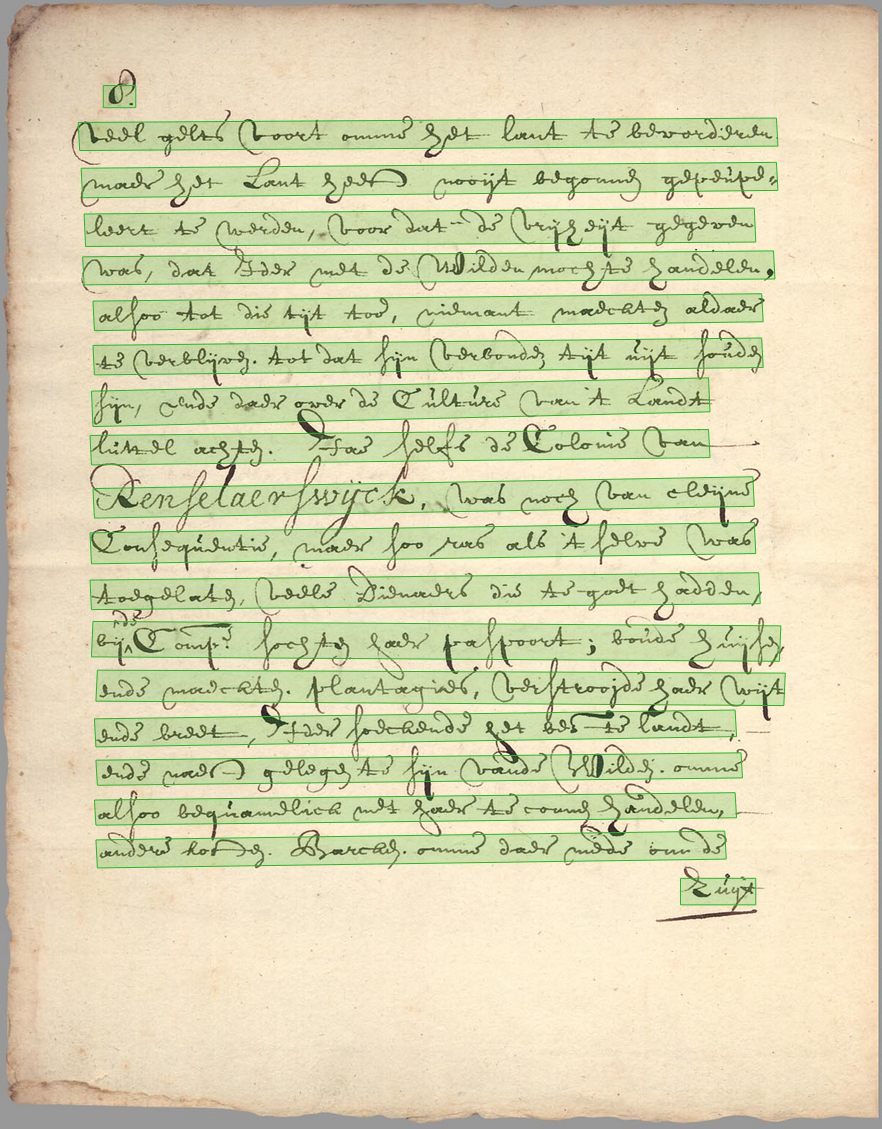}
        \includegraphics[width=0.2\textwidth]{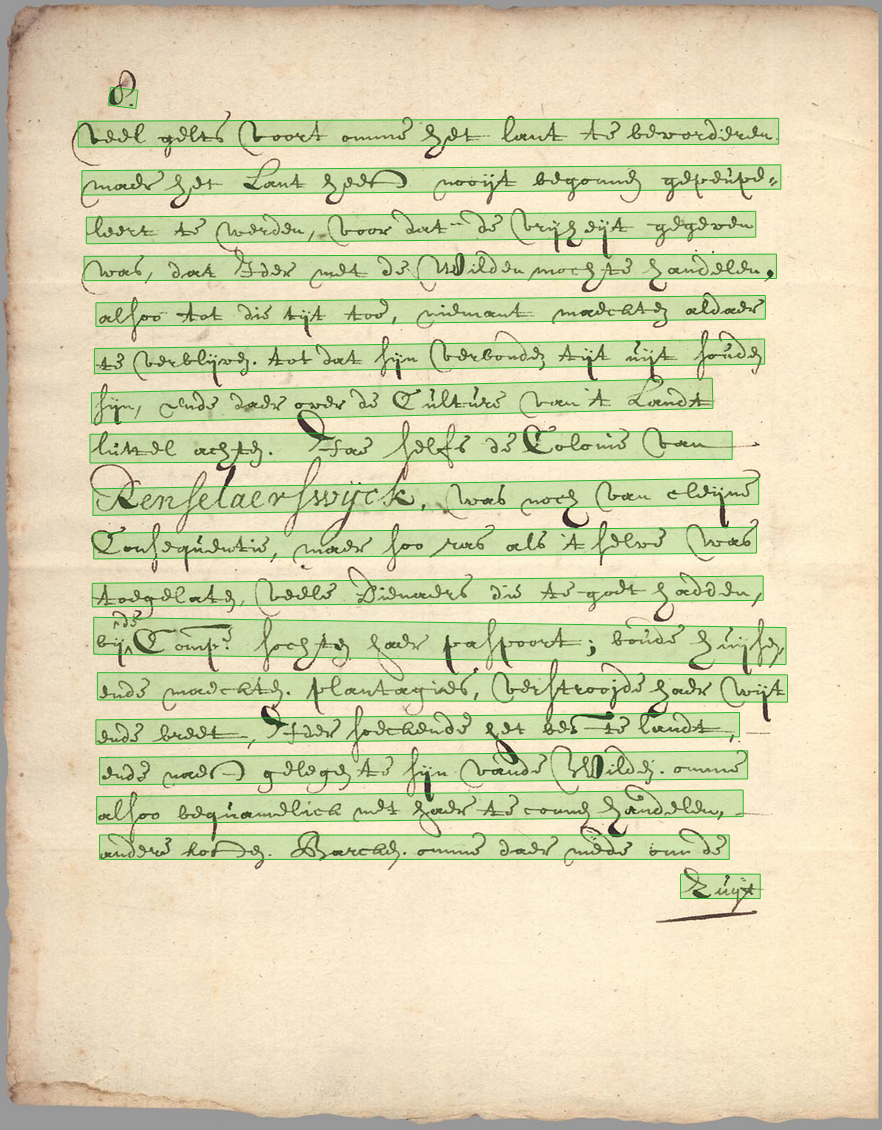} \\
        \vspace*{0.2cm}
        \includegraphics[width=0.2\textwidth]{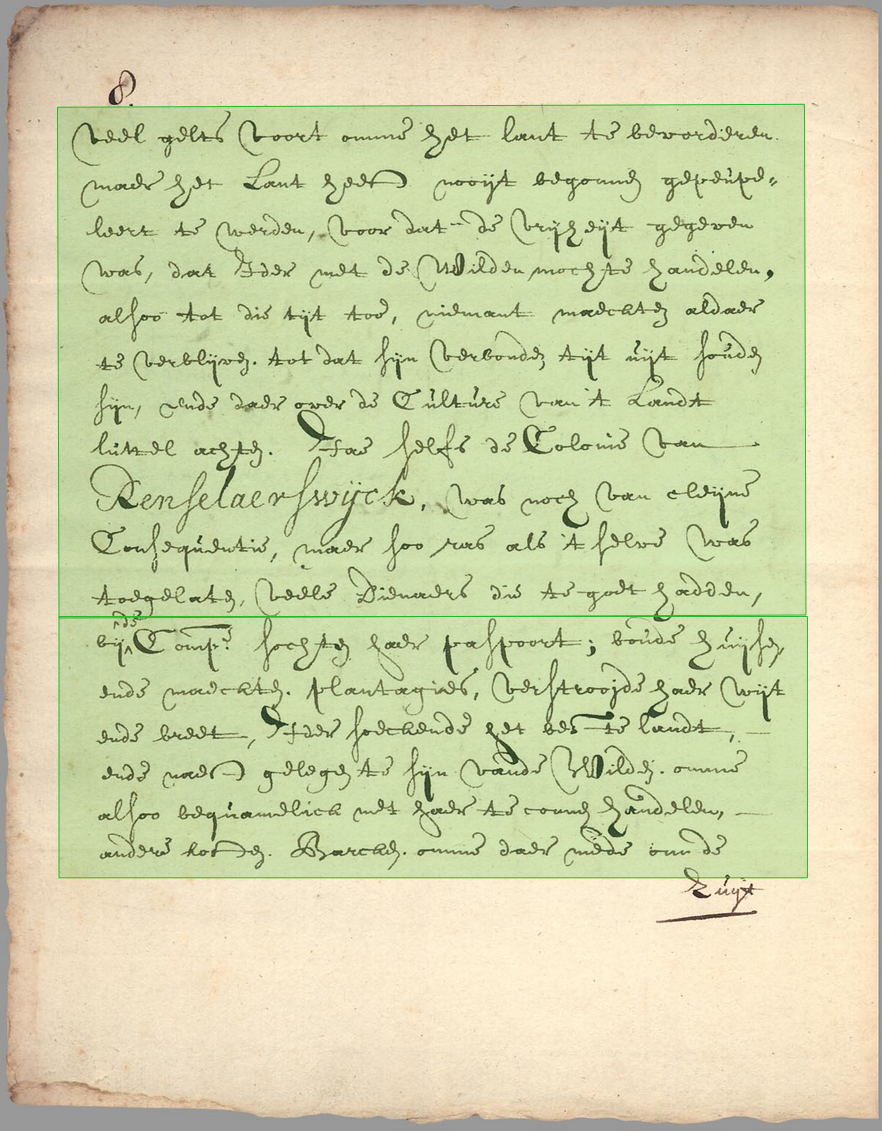}
        \includegraphics[width=0.2\textwidth]{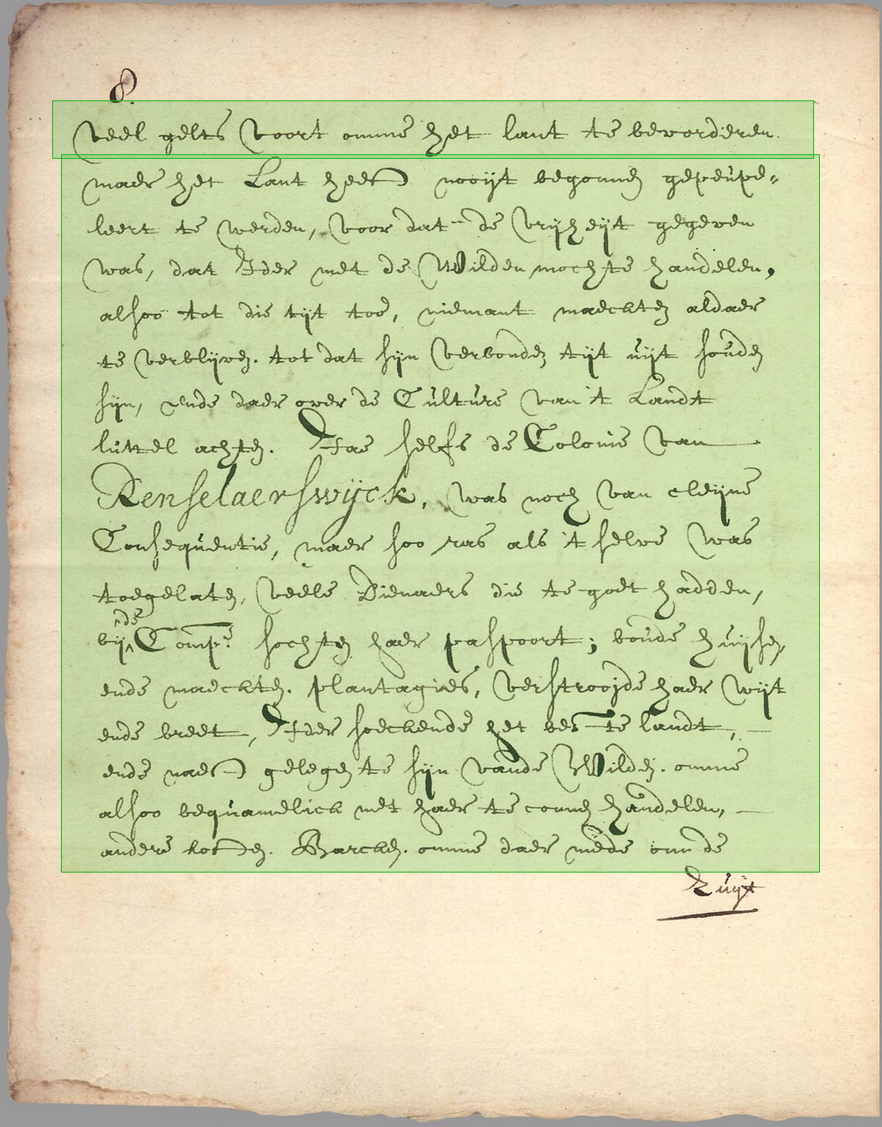}
    \end{center}
    \caption{Predictions made by generic (top) and specific (bottom) models on a ScribbleLens page. Left images show the results obtained by Doc-UFCN and the right ones by dhSegment.}
    \label{fig:scribblelens_home-cartularies_res1}
\end{figure}

\begin{figure}[t]
    \begin{center}
        \includegraphics[width=0.2\textwidth]{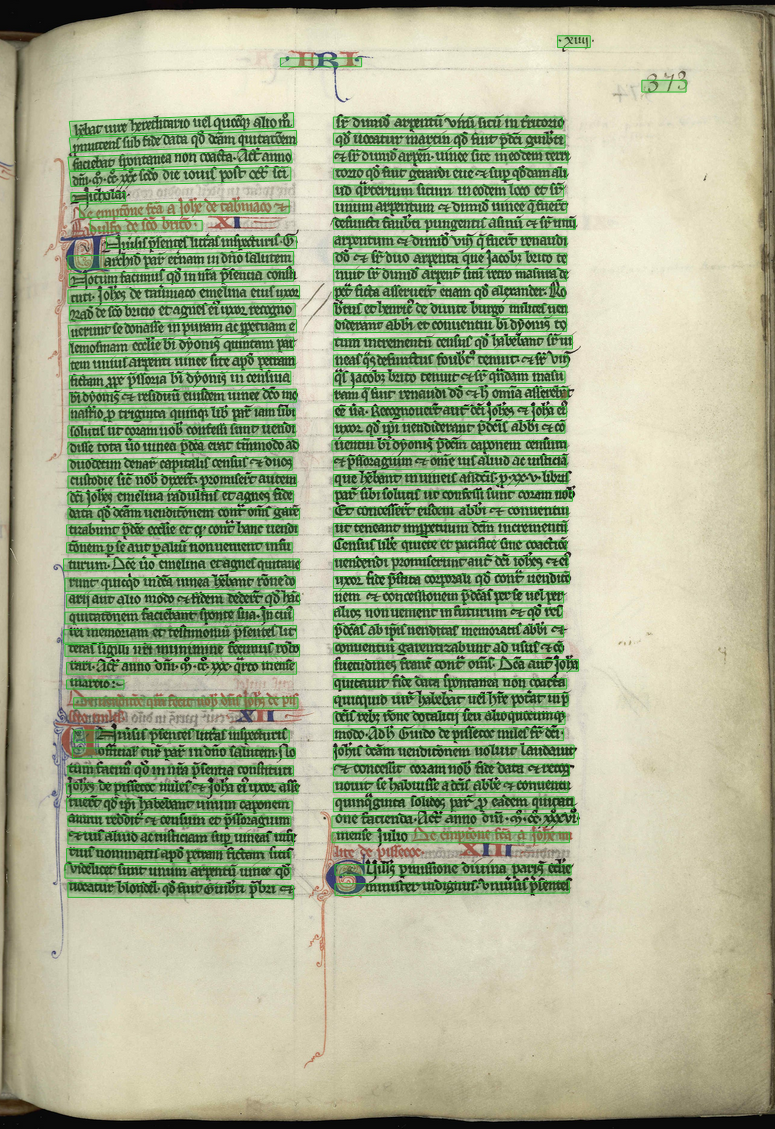}
        \includegraphics[width=0.2\textwidth]{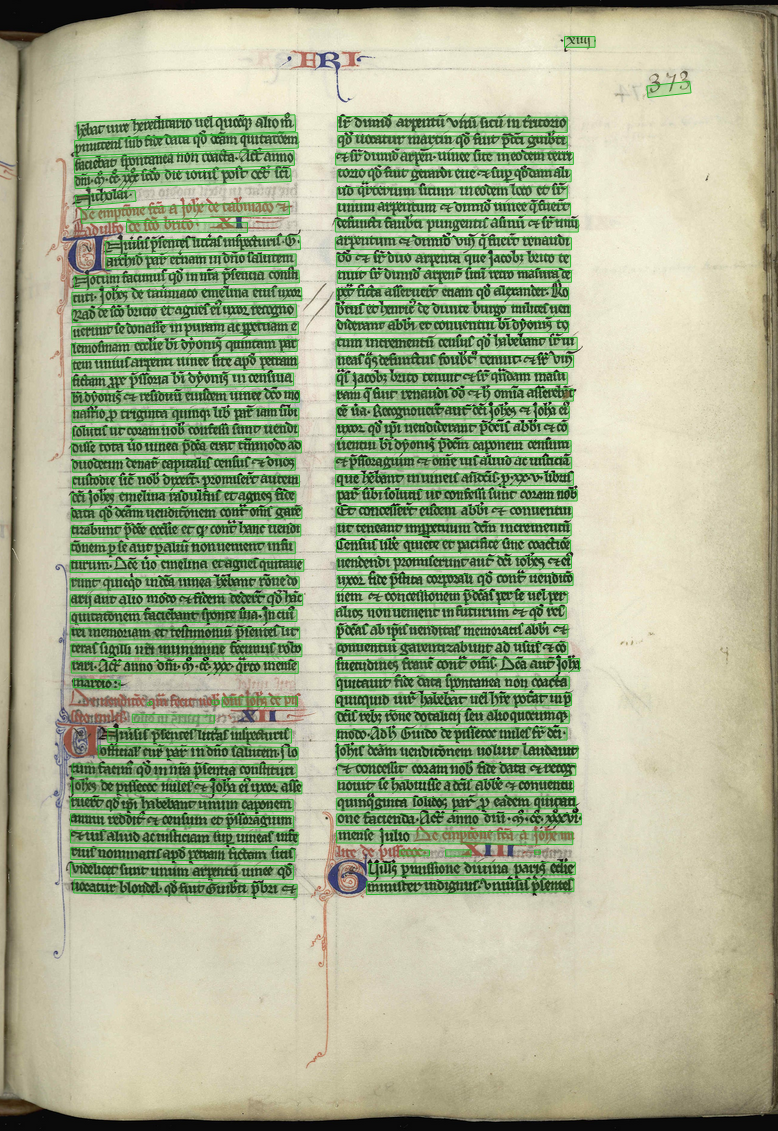}
    \end{center}
    \caption{Predictions made by the generic Doc-UFCN (left) and dhSegment (right) models on a page from Alcar dataset.}
    \label{fig:scribblelens_home-cartularies_res2}
\end{figure}

The small AP scores of the generic models can be explained by the way the dataset has been annotated: really high bounding polygons. The models having been trained on well split and way thinner polygons, only a few ground-truth polygons have been paired to predicted ones during the AP computation. The same observation applies to Alcar results. Figures \ref{fig:scribblelens_home-cartularies_res1} and \ref{fig:scribblelens_home-cartularies_res2} show a visualization of the results obtained on ScribbleLens and Alcar images. Despite bad metric values, the generic models seem to clearly outperform specific models, hence the importance of developing generic models.

\subsubsection{Impact of the annotations unification}

Unsurprisingly, according to Table \ref{tab:res_obj} almost all the values are better when using the model trained on the unified labels, sometimes by a considerable margin (+\textit{33} percentage points for Balsac and +\textit{37} for Bozen). For DIVA-HisDB dataset, the results are mixed. We assume that this is due to the unification process that can alter substantially the labels by reducing the line height. \\

These object-level metrics have underlined their necessity to be used along with pixel-level values to evaluate and compare models. However, it is still hard to see the benefit of using generic models on unseen documents. Goal-directed metrics, described in the next section, will allow a further better comparison of predicted and ground-truth objects.

\section{Goal-oriented evaluation}
\label{sec:gd_eval}
In the previous sections, we have discussed the results at pixel and object-levels for the three methods. We have shown that the object metrics provide more information about the quality and performances of a line segmentation model than pixel-level measures can provide. However, they still face limitations when the predicted objects are too thin compared to the ground-truth ones. Goal-directed measures can help determine the actual capacities of the models when facing this.

We conducted a goal-oriented evaluation of the segmentation systems on the five datasets for which the  ground-truth of the text line transcription is available using Character Error Rate (CER) and Word Error Rate (WER). For the sake of clarity, in the following tables, only the CER are presented since WER are strongly related to CER.

To conduct this goal-oriented evaluation, we used a Handwritten Text Recognizer \cite{boros2020} based on the KALDI library \cite{arora2019}. The model is composed of two main components: an optical model using a hybrid Deep Neural Network-Hidden Markov Model and a language model based on a \textit{n}-gram model trained on subwords generated by Byte Pair Encoding method. Note that contrary to the line segmentation model which is trained on all the datasets, we trained one specific model for each dataset and used this model for the goal-oriented evaluation.

The next paragraphs present and analyze the segmentation results using two metrics based on the CER at page and line-levels.

\subsection{CER at page-level}

\begin{figure}
\vspace{-1cm}
    \begin{center}
        \def\svgwidth{0.42\textwidth}
        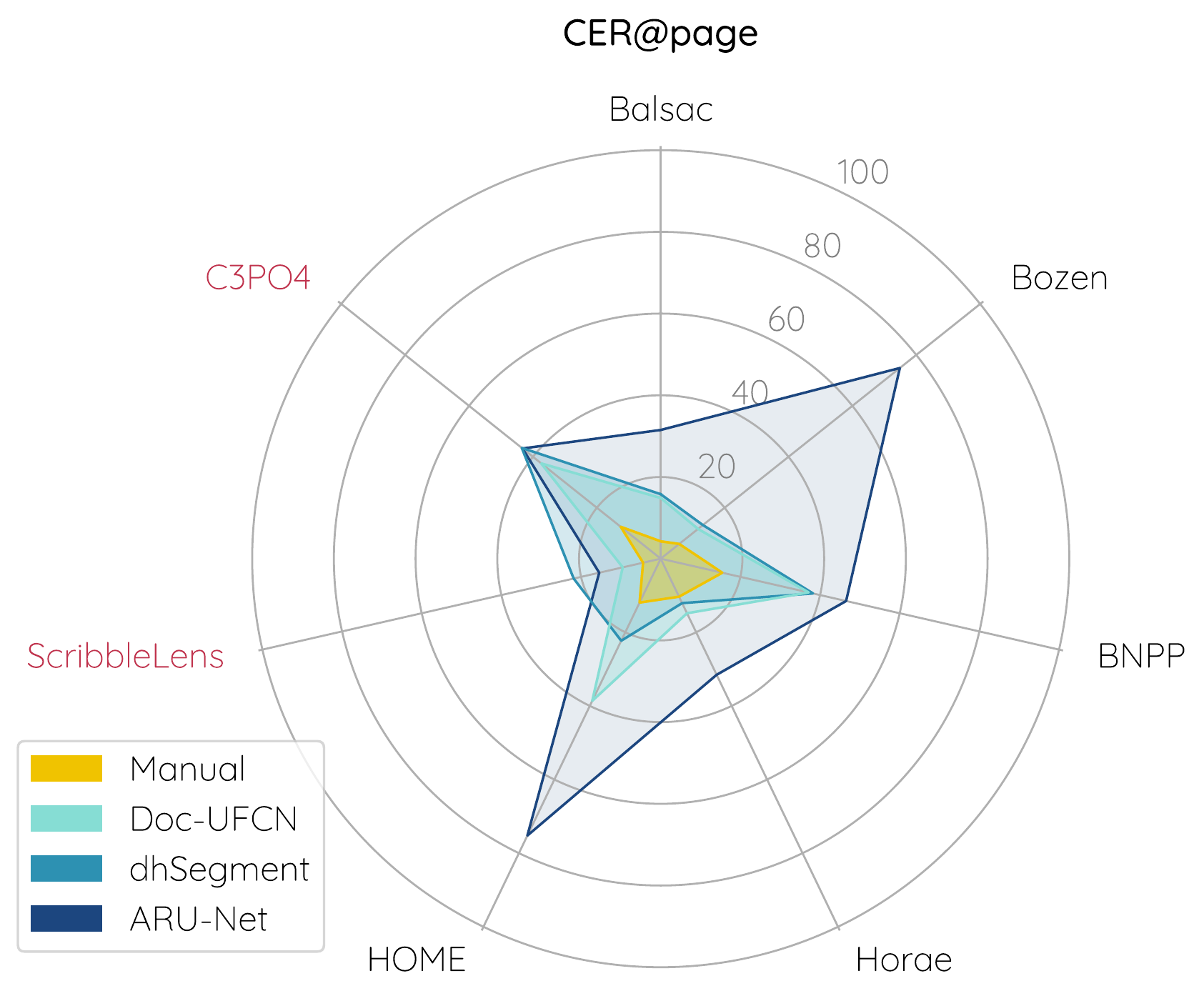
    \end{center}
    \caption{CER at page-level on the test sets.}
    \label{fig:res_cer_page}
\end{figure}

To start the evaluation, we first chose to compute the CER at page-level. All the predicted and ground-truth line polygons of an image are sorted from top-left to bottom-right. Following this order, all transcriptions are concatenated in a single line of text and the CER@ \\ page is computed. Table \ref{tab:res_cer_page} and Figure \ref{fig:res_cer_page} show the CER@page obtained by the systems. In addition, we computed the CER obtained by the HTR system when transcribing the ground-truth line polygons. Therefore, the \textit{Manual} column in the following Tables corresponds to the best achievable CER with the ideal segmentation system. It means the CER we would have if we had \textit{100}\% for all the pixel and object-level metrics.

\subsubsection{Systems comparison on the training datasets}

Results in Table \ref{tab:res_cer_page} show the low performance of the HTR on the lines detected by ARU-Net. This high error rate is the consequence of many merged detected text lines which cannot be correctly recognized,  as it was already highlighted with the object-level evaluation.

Doc-UFCN performs better than dhSegment on three of the five datasets, and it is slightly behind dhSegment for the Horae dataset. These results confirm those obtained with the pixel and object-level evaluations. Doc-UFCN is however far behind dhSegment on the HOME dataset, contrarily to the results obtained using pixel and object metrics.

If we analyze more in depth the segmentation results obtained by Doc-UFCN on the HOME dataset, we see that about half of the pages have been perfectly segmented without any mergers, significantly raising the AP scores. However, the other pages contain predicted lines that are mergers of two, three lines, or even mergers of lines of whole paragraphs. This leads to slightly descreasing the AP scores but drastically degrading the CER performances. Indeed, while one single merger has only a low impact on the AP score, it directly impacts the CER by two types of errors:
\begin{itemize}
    \item The CER between the prediction and its corresponding ground-truth line (which is often high in case of a merger);
    \item The CER of the non-matched ground-truth lines, which is equal to the length of each non-matched line.
\end{itemize}
This second error is not significant when only a few ground-truth lines are not paired. This is the case with the first four datasets, where the number of mergers is negligible. It is even less significant when the non-paired lines have a small number of characters. However, HOME is the dataset with the largest number of characters per line (up to \textit{6} times larger than the other datasets). That is why this second error really impacts the final CER on HOME dataset. This is also the reason the AP scores do not reveal the phenomenon.

\begin{figure*}
    \begin{center}
        \def\svgwidth{0.9\textwidth}
        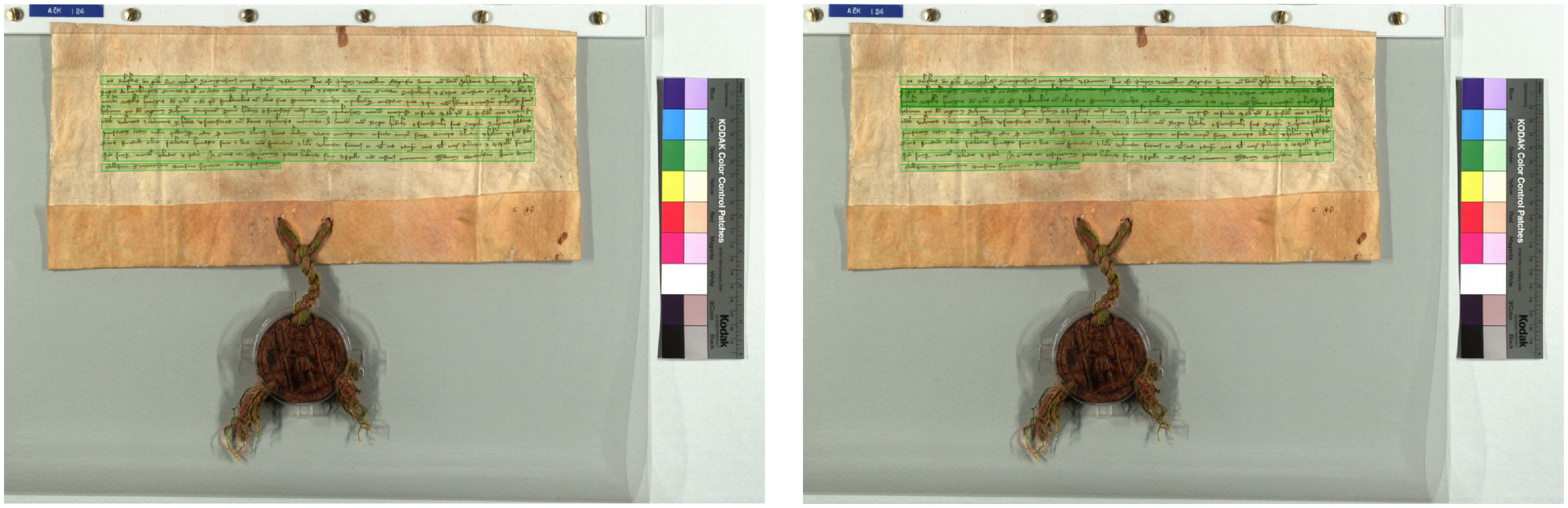 
    \end{center}
    \caption{Simulation of the results when two lines are well split (left) and merged (right) on a page from HOME dataset. On the left, mean AP[.5, .95]=\textit{60}\% and CER@page=\textit{7.3}; on the right mean AP[.5, .95]=\textit{51.3}\% and CER@page=\textit{20.4}.}
    \label{fig:res_home_visualization}
\end{figure*}

Figure \ref{fig:res_home_visualization} illustrates this point: the left image is correctly segmented while on the right one, two lines are merged. In this case, introducing a merger in the predictions leads to a decrease in the mean AP[.5, .95] of \textit{15}\% (\textit{60}\% to \textit{51.3}\%) while the CER@page degraded by \textit{179}\% (\textit{7.3} to \textit{20.4}). This proves that a merger does not affect the different metrics the same way. \\

On the HOME dataset, unlike Doc-UFCN, dhSegment shows a less accurate localization of the text lines (lower AP scores) but very few mergers, leading to better recognition performances. Indeed, HOME is the dataset where the handwriting is the densest and the lines are the closest to each other among the 10 datasets. Due to the really small height of the text lines and the resizing to \textit{768} pixels, we think that Doc-UFCN is not the best adapted architecture to work with these pages, contrarily to dhSegment that is doing better since it works at the original image size. \\

This additional metric gives again more insight on the models performances, being complementary to previously seen metrics. It can indeed detect behaviors not highlighted by pixel nor object measures.

\begin{table}
    \setlength\tabcolsep{3pt}
    \caption{Results of the HTR at page-level on the testing sets with the labels unification process obtained by Doc-UFCN (UFCN), dhSegment (dhSeg) and ARU-Net. UFCN \xmark{} shows the results of Doc-UFCN when trained on original non-normalized labels. ScribbleLens* reports the results of the specific models.}
    \label{tab:res_cer_page}
    \begin{center}
        \begin{tabular}{l|r|rrrr}
            \hline\noalign{\smallskip}
            \multicolumn{1}{c}{\multirow{2}{*}{\textsc{Dataset}}} & \multicolumn{5}{c}{CER@page} \\
            \multicolumn{1}{c}{} & \multicolumn{1}{c}{Manual} & \multicolumn{1}{c}{UFCN} & dhSeg & ARU & UFCN \xmark{} \\ \noalign{\smallskip}\hline\noalign{\smallskip}
            Balsac & 4.3 & 14.9 & 15.8 & 31.5 & \color{best_result}\underline{14.4} \\
            BNPP & 15.5 & 37.2 & 38.2 & 46.5 & \color{best_result}\underline{34.4} \\
            Bozen & 5.8 & \color{best_result}\underline{11.7} & 13.2 & 74.9 & 27.6 \\
            HOME & 11.9 & 38.6 & \color{best_result}\underline{22.3} & 75.2 & 33.5 \\
            Horae & 10.3 & 14.8 & \color{best_result}\underline{12.1} & 31.5 & 15.1 \\
            \textbf{mean} & N/A & \textbf{23.4} & \textbf{\color{bf_best_result}\underline{20.3}} & \textbf{51.9} & \textbf{25.0} \\ 
            \noalign{\smallskip}
            \hline\noalign{\smallskip}
            ScribbleLens & 4.4 & \color{best_result}\underline{9.5} & 21.9 & 15.4 & 12.9 \\
            Alcar & 12.5 & \color{best_result}\underline{37.4} & 43.5 & 43.3 & 43.2 \\ \noalign{\smallskip}
            \hline\noalign{\smallskip}
            ScribbleLens* & 4.4 & \color{best_result}\underline{25.2} & 92.6 & -~~ & -~~ \\\noalign{\smallskip}
            \hline
        \end{tabular}
    \end{center}
\end{table}

\subsubsection{Generalization to unseen datasets}

Generalization results are also presented in Table \ref{tab:res_cer_page}. For ScribbleLens, we see the advantage of using a generic model: the results of specific models (Doc-UFCN \textit{25.2}\% CER, dhSegment \textit{92.5}\% CER) are dramatically worse than those of the generic ones (Doc-UFCN \textit{9.5}\% CER, dhSegment \textit{21.9}\% CER). The CER values of Alcar are large for all the systems, which can be due to the complexity of some document images:  bad quality of the scanning, poor preservation conditions (for example, some pages have been torn). However, these results still highlight the generalization capacities of the generic Doc-UFCN model, yielding better results on ScribbleLens than the specific model.

\subsubsection{Impact of the annotations unification}

The two Doc-UFCN with and without the unification process have quite similar results on four datasets without any significant degradation. However, the impact of the label normalization is more important on the  Bozen database. For the same reason as ARU-Net, the model predicts a lot of merged lines, leading to poor HTR results compared to the model trained on the non-normalized ground-truth. The unification process did not improve the results on Balsac and BNPP because the original annotations were already thin and a correct input for the HTR system. 

Even if training with the normalized ground-truth did not show a significant improvement on the CER values for four datasets, it really impacted the Bozen predictions. Regarding the unseen datasets, the annotations unification has also a positive impact. The model trained with the normalized ground-truth yields a CER of \textit{9.5\%} for ScribbleLens and \textit{37.4\%} for Alcar which respectively corresponds to \textit{26\%} and \textit{13\%} of relative error decrease.

\begin{table*}
    \setlength\tabcolsep{3.2pt}
    \caption{Results of the HTR system at line-level on the testing sets with the labels unification process obtained by Doc-UFCN (UFCN), dhSegment (dhSeg) and ARU-Net. The second part of the table shows the quantity of predicted and annotation characters matched before computing the CER. UFCN \xmark{} shows the results of Doc-UFCN when trained on original non-normalized labels. ScribbleLens* reports the results of the specific models.}
    \label{tab:res_cer_line}
    \begin{center}
        \begin{tabular}{l|r|rrrr|rrrr|rrrr}
            \hline\noalign{\smallskip}
            \multicolumn{1}{c}{\multirow{2}{*}{\textsc{Dataset}}} & \multicolumn{1}{c}{CER} & \multicolumn{4}{c}{\textsc{CER@.5}} & \multicolumn{4}{c}{\textsc{CER@.75}} & \multicolumn{4}{c}{\textsc{CER@[.5, .95]}} \\
            \multicolumn{1}{c}{} & \multicolumn{1}{c}{Manual} & \multicolumn{1}{c}{UFCN} & dhSeg & ARU & \multicolumn{1}{c}{UFCN \xmark{}} & UFCN & dhSeg & ARU & \multicolumn{1}{c}{UFCN \xmark{}} & UFCN & dhSeg & ARU & UFCN \xmark{} \\ \noalign{\smallskip}\hline\noalign{\smallskip}
            Balsac & 4.3 & \color{best_result}\underline{7.2} & 8.2 & 29.7 & \color{best_result}\underline{7.2} & 6.4 & 8.4 & 41.0 & \color{best_result}\underline{6.3} & \color{best_result}\underline{14.2} & 14.9 & 52.0 & 17.1 \\
            BNPP & 15.6 & 22.4 & 21.5 & 32.1 & \color{best_result}\underline{18.8} & 28.0 & 28.8 & 41.4 & \color{best_result}\underline{22.5} & 44.2 & 42.8 & 53.2 & \color{best_result}\underline{36.0} \\
            Bozen & 5.8 & \color{best_result}\underline{8.8} & 10.0 & 86.5 & 27.3 & \color{best_result}\underline{8.6} & 9.80 & 94.0 & 34.8 & 20.7 & \color{best_result}\underline{18.2} & 93.3 & 47.3 \\
            HOME & 12.0 & 36.1 & \color{best_result}\underline{23.3} & 80.1 & 29.4 & 60.0 & \color{best_result}\underline{29.5} & 94.6 & 36.2 & 61.8 & \color{best_result}\underline{42.0} & 91.1 & 46.6 \\
            Horae & 10.6 & 15.2 & \color{best_result}\underline{12.0} & 30.3 & 15.7 & 15.0 & 17.5 & 49.6 & \color{best_result}\underline{12.9} & 22.6 & 29.0 & 58.0 & \color{best_result}\underline{20.6} \\
            \noalign{\smallskip}
            \hline\noalign{\smallskip}
            ScribbleLens & 4.6 & \color{best_result}\underline{9.8} & 18.2 & 15.8 & 14.3 & \color{best_result}\underline{26.2} & 77.0 & 34.0 & 58.0 & \color{best_result}\underline{40.3} & 60.0 & 45.5 & 54.2 \\
            Alcar & 12.5 & \color{best_result}\underline{22.9} & 27.6 & 30.0 & 31.7 & \color{best_result}\underline{22.2} & 42.2 & 40.4 & 58.4 & \color{best_result}\underline{46.6} & 49.3 & 54.0 & 61.3 \\ \noalign{\smallskip}
            \hline\noalign{\smallskip}
            ScribbleLens* & 4.6 & \color{best_result}\underline{24.3} & 90.9 & -~~ & -~~ & \color{best_result}\underline{23.5} & 93.1 & -~~ & -~~ & \color{best_result}\underline{32.6} & 93.3 & -~~ & -~~ \\ \noalign{\smallskip}
            \hline\noalign{\smallskip}
            \multicolumn{14}{c}{Amount of characters of the ground-truth lines that are matched with a prediction line.} \\
            \multicolumn{14}{c}{1 means that 100\% of the ground-truth characters are matched to a predicted line.} \\ \noalign{\smallskip}
            \hline\noalign{\smallskip}
            Balsac & & 0.95 & 0.95 & 0.64 & \color{best_result}\underline{0.97} & 0.84 & 0.73 & 0.33 & \color{best_result}\underline{0.86} & & \\
            BNPP & & \color{best_result}\underline{0.93} & 0.83 & 0.73 & 0.87 & 0.29 & 0.32 & 0.30 & \color{best_result}\underline{0.45} & & \\
            Bozen &  & \color{best_result}\underline{0.94} & 0.93 & 0.12 & 0.67 & 0.73 & \color{best_result}\underline{0.79} & 0.04 & 0.43 & \\
            HOME & & 0.61 & \color{best_result}\underline{0.79} & 0.18 & 0.73 & 0.20 & \color{best_result}\underline{0.47} & 0.04 & 0.47 & & \\
            Horae & 100\% & \color{best_result}\underline{0.98} & 0.97 & 0.90 & 0.98 & \color{best_result}\underline{0.83} & 0.53 & 0.32 & 0.78 & \multicolumn{4}{c}{N/A} \\
            \noalign{\smallskip}
            \cline{1-1}\cline{3-10}\noalign{\smallskip}
            ScribbleLens & & 0.80 & 0.37 & \color{best_result}\underline{0.81} & 0.71 & 0.14 & 0.02 & \color{best_result}\underline{0.23} & 0.06 & \\
            Alcar & & \color{best_result}\underline{0.92} & 0.73 & 0.72 & 0.79 & 0.28 & 0.18 & \color{best_result}\underline{0.38} & 0.12 \\ \noalign{\smallskip}
            \cline{1-1}\cline{3-10}\noalign{\smallskip}
            ScribbleLens* & & \color{best_result}\underline{0.76} & 0.10 & -~~ & -~~ & \color{best_result}\underline{0.71} & 0.07 & -~~ & -~~ & \\ \noalign{\smallskip}
            \hline
        \end{tabular}
    \end{center}
\end{table*}

\subsection{CER at line-level}

This last metric is closely linked to the CER at page-level. The CER here is not computed on a large line of text representing the whole page, but on each single predicted text line. In this respect, the predicted and ground-truth lines must first be paired. In the literature, they are often paired based on a IoU threshold of \textit{t} = \textit{50}\%. As for the AP, we computed the CER for two IoU thresholds of \textit{50}\% (CER@.5) and \textit{75}\% (CER@.75) as well as an average over the range \textit{50}\%-\textit{95}\% of IoU (CER@[.5, .95]). The predicted lines are paired with the ground-truth ones that have the highest IoU, such that only one prediction can be paired with an annotation and conversely. Once the lines are paired, we compute the CER for all the couples with an IoU higher than the threshold. In addition, the final CER is penalized by all the lines that are not paired.

Table \ref{tab:res_cer_line} presents the results obtained after the HTR system at line-level. The second part of the table shows the percentage of ground-truth characters that have been matched to a predicted line to compute the CER values. One could have done this at line-level (percentage of matched lines) to see on what quantity of lines the CER have been computed. However, since lines can contain a variable number of characters, it wouldn't precisely reflect the real amount of matchings.

As for the previous metrics, ARU-Net is not competitive: not enough matched characters and really high CER values. To compare Doc-UFCN and dhSegment, one needs to check the CER and matching percentage as a whole. Indeed, having a really low CER when computed only on a small part of the predicted lines is not meaningful, since some lines can be easier to recognize. It is preferable to have a good compromise between the number of matched characters and the error rate.

The results obtained at line-level really reflect the ones obtained with the AP scores and CER at page-level. At \textit{50}\% of IoU, Doc-UFCN seems better for Balsac and Bozen datasets. When being less permissive with a IoU of \textit{75}\%, this trend continues for Balsac and starts to reverse for Horae. In addition, it becomes harder to determine the best model for BNPP and Bozen datasets, both models yielding equivalent results. Indeed, when looking at the averaged results CER@[.5, .95], both dhSegment and Doc-UFCN have similar performances. With the aim of having a generic historical model, both architectures seem appropriate, yielding good results at pixel and object-levels and acceptable character error rates at page and line-levels.

In previous sections, we have shown that using pixel metrics like the IoU for evaluating models is not sufficient, since it does not reflect the quality of the predicted objects. We presented object-level metrics and showed that they are complementary to pixel-level measures to have more information about a model performance and a more accurate comparison of systems. We believe that these Average Precision measures should be used as much as possible to fairly evaluate models. We also introduced goal-directed metrics and exposed that, when one has transcriptions, they can give additional details on the predicted objects and give a first idea on how a HTR system would behave.

\section{Conclusion}
\label{sec:conclusion}
In this paper, we have shown that training a generic model for text line segmentation in historical documents is possible. We trained three state-of-the-art models, achieving good performance on various datasets. This was made possible by creating a large and diverse training set, which is, to the best of our knowledge, the biggest and most diversified historical dataset used for comparing text segmentation systems. We also showed that, when aggregating different datasets, normalizing the ground-truth of the annotated bounding polygons reduces the labels inconsistencies among the annotated corpora and allows training better models. In addition, the generic models trained on multiple datasets can be better both on each individual dataset and on documents unseen during the training phase, proving their generalization ability.

For a fair performance evaluation of the three systems, this paper also compares and analyses several segmentation metrics. We showed that the pixel-level metrics are not sufficient, as they do not account for the quality of the predicted objects. To overcome this drawback, line-level metrics have been introduced. They have shown that ARU-Net is not appropriate enough for the line segmentation task when trained with such annotations, the amount of line mergers being not negligible compared to the two other approaches. This system is indeed often used to detect baselines from the documents, that are thinner and more spaced than line bounding polygons. These measures also confirmed the good performances of Doc-UFCN and dhSegment on most of the datasets, providing precise and accurate object detection. These findings would not have been possible using pixel-level metrics only. We strongly believe that using Average Precision scores is necessary to correctly evaluate line segmentation models. Our evaluation library will be made publicly available\footnote{\url{https://gitlab.com/teklia/dla/document_image_segmentation_scoring}}. Based on the annotations, it can be used on any dataset.

Lastly, this paper goes further by providing a goal-directed evaluation which, to our knowledge, has never been conducted so far on this task. The HTR evaluation metrics give even more information on the predicted objects, being complementary to object-level metrics. In addition, they allow exploring the impact of the segmented lines quality on the final recognition results.

Our future works will focus on building generic models with more diverse training data. Besides, we aim to run the line segmentation in a semi-supervised way using the introduced HTR metrics.

\section*{Acknowledgments}
\sloppy
This work is part of the \textit{HOME History of Medieval Europe} research project, supported by the European JPI Cultural Heritage and Global Change (Grant agreements No. ANR-17-JPCH-0006 for France). It is also supported by the i-BALSAC project with the Université du Québec in Chicoutimi and benefited from the support of the project HORAE ANR-17-CE38-0008 of the French National Research Agency (ANR). Mélodie Boillet is partly funded by the CIFRE ANRT grant No. 2020/0390. Lastly, this work was granted access to the HPC resources of IDRIS under the allocation 2021-AD011011910 made by GENCI.

\fussy

\bibliographystyle{unsrt}
\bibliography{references}

\end{document}